\documentclass{article}

     \PassOptionsToPackage{square,sort,comma,numbers}{natbib}

\usepackage[preprint]{neurips_2019}

\usepackage[utf8]{inputenc} 
\usepackage[T1]{fontenc}    
\usepackage{hyperref}       
\usepackage{url}            
\usepackage{booktabs}       
\usepackage{amsfonts}       
\usepackage{nicefrac}       
\usepackage{microtype}      
\usepackage{mwe}    
\usepackage{subfig}
\usepackage{multirow}
\usepackage{adjustbox}
\usepackage{color}
\usepackage{xcolor}
\usepackage{wrapfig}
\usepackage{graphicx,wrapfig}
\usepackage{gensymb}
\usepackage{tikz}
\usepackage{caption}
\usepackage{mathtools}
\usepackage{amsfonts, amsmath}
\usepackage{wrapfig}
\usepackage{algorithm}%
\usepackage[noend]{algpseudocode}
\usepackage{enumitem}
\usepackage{comment}

\title{Learning Latent Dynamics for Partially-Observed Chaotic Systems}

\author{%
  Said Ouala$^1$, Duong Nguyen$^1$, Lucas Drumetz$^1$, Bertrand Chapron$^2$ \\
  \textbf{Ananda Pascual}$^3$, \textbf{Fabrice Collard}$^4$, \textbf{Lucile Gaultier}$^4$, \textbf{Ronan Fablet}$^1$\\
   $(1)$ IMT Atlantique; Lab-STICC, Brest, France\\
   $(2)$ Ifremer, LOPS, Brest, France\\
   $(3)$ IMEDEA, UIB-CSIC, Esporles, Spain\\
   $(4)$ ODL, Brest, France\\
  \texttt{\{said.ouala, van.nguyen1, lucas.drumetz, ronan.fablet\}@imt-atlantique.fr}\\
  \texttt{Bertrand.Chapron@ifremer.fr}, \texttt{ananda.pascual@imedea.uib-csic.es} \\
  \texttt{\{dr.fab, lucile.gaultier\}@oceandatalab.com}
}

\begin{document}

\maketitle

\begin{abstract}
This paper addresses the data-driven identification of latent dynamical representations of partially-observed systems, {\em i.e.} dynamical systems for which some components are never observed, with an emphasis on forecasting applications, including long-term asymptotic patterns. Whereas state-of-the-art data-driven approaches rely on delay embeddings and linear decompositions of the underlying operators, we introduce a framework based on the data-driven identification of an augmented state-space model using a neural-network-based representation. For a given training dataset, it amounts to jointly learn an ODE (Ordinary Differential Equation) representation in the latent space and reconstructing latent states. Through numerical experiments, we demonstrate the relevance of the proposed framework w.r.t. state-of-the-art approaches in terms of short-term forecasting performance and long-term behaviour. We further discuss how the proposed framework relates to Koopman operator theory and Takens' embedding theorem.
 \end{abstract}

\section{Introduction}
Learning the underlying dynamical representation of some observed variables ${\mathrm{x}}_{t} \in \mathbb{R}^n$, where $t \in \left\{t_0,...,T\right\}$ is the temporal sampling and $n$ the dimension of our observation space is a key challenge in various scientific fields, including control theory, geoscience, fluid dynamics, economics; for applications ranging from system identification to forecasting and assimilation issues \cite{ident_control,Abarbanel1996,jeong_hussain_1995,iden_economics}.

For fully-observed systems, {\em i.e.} when observed variables ${\mathrm{x}}_{t}$ relate to some underlying deterministic states ${\mathrm{z}}_{t}$ according to a diffeomorphic mapping possibly corrupted by some noise, recent advances \cite{brunton_discovering_2016,fablet_blin_ieee,nguyen2019emlike} have shown that one can identify the governing equations of the dynamics of ${\mathrm{z}}$ from a representative dataset of observations $\{{\mathrm{x}}_{t_i}\}_i$. When the observed variables ${\mathrm{x}}_{t}$ only relate to some but not all the components of underlying states ${\mathrm{z}}_{t}$, these approaches do not apply since no ODE or, more generally, no one-to-one mapping defined in the observation space can represent the time evolution of the observations. In this context, Takens's theorem states the conditions under which a delay embedding, formed by lagged versions of the observed variables, guarantees the existence of governing equations in the embedded space \cite{takens_theorem}. 

Takens's theorem has motivated a rich literature of machine learning schemes to identify dynamical representations of partially-observed systems using a delay embedding. This comprises both non-parametric schemes based on nearest-neighbors or analogs \cite{Abarbanel_model_chaos} as well as parametric schemes which include polynomial representations \cite{paduart_identification_2010}, neural network models \cite{NN_takens}, Support Vector Regression (SVR) models \cite{SVM_takens}. For all these approaches, the identification of the appropriate delay embedding is a critical issue \cite{takens_params_1,takens_params_2}.


In this work, we do not rely explicitly on a delay embedding. We rather aim to identify an augmented space of higher dimension than the manifold spanned by the observed variables, where the dynamics of the observations can be fully described by an ODE. Numerically, the joint identification of the augmented space and the governing ODE for a given observation dataset exploits neural-network representations. We report experiments for linear and chaotic dynamics, which illustrate the relevance of the proposed framework compared to state-of-the-art approaches, and then further discuss the key features of this framework with respect to state-of-the-art dynamical systems identification tools such as Koopman operator theory \cite{KoopmanGeneral}.


\section{Background and Related Work}
\label{sec:related_works}
Let us consider an unobserved state variable $\boldsymbol{\mathrm{z}}$ governed by an autonomous system of $s$ differential equations $\Dot{{\mathrm{z}}}_{t} = f({\mathrm{z}}_{t})$. Let us also assume that this system generates a flow $\Phi_{t_i}({\mathrm{z}}_{0}) = \int_{t_0}^{t_i}f({\mathrm{z}}_{u})du \in \mathbb{R}^s$ with trajectories that are characterized by an asymptotic limit-cycle $L$ of dimension $d$ contained in $\mathbb{R}^s$. We further assume that we are provided with a measurement function $\mathcal{H}$ that maps our state variable $\mathrm{z}$ to our observation space $\mathrm{x}_t = \mathcal{H}(\mathrm{z}_t) \in \mathbb{R}^n$.

When considering the data-driven identification of a dynamical mapping that governs some observation data, we first need to evaluate if the dynamics expressed in our observation space are deterministic and thus, can be described using a smooth\footnote{The word smooth here stands for continuously differentiable or $\mathcal{C}^1$.} ODE. Another way to tackle this question is to find the conditions under which the deterministic properties of the unobserved limit-cycle $L$ are preserved in the observation space in $\mathbb{R}^n$ so one can reliably perform forecasts in the observation space.

The general condition under which a mapping $\mathcal{H}$ will preserve the topological properties of the initial limit-cycle involves a differential structure. Assuming that $L$ is a smooth compact differential manifold, the topological properties of $L$ are preserved through a projection $\mathcal{H}$ in $\mathbb{R}^n$ if $\mathcal{H}$ is one-to-one and an immersion of $L$ in $\mathbb{R}^n$. Under these conditions our observation mapping is called an embedding \cite{Sauer1991}.

The simplest example of an embedding is the case where our observation operator is an identity matrix. With such embedding we have direct access to the state variable $z$ which is governed by a deterministic ODE. This particular case has been widely studied in the literature, parametric representations based on augmented state formulations have been for decades the most popular models due to their simplicity and interpretability \cite{paduart_identification_2010}, \cite{brunton_discovering_2016}. In the last years, neural network and deep learning have enriched the state-of-the-art of parametric representations \cite{wiewel2018latentspace}, \cite{raissi2018multistep}. In particular, the link between residual networks \cite{chen2018neural}, \cite{ResINN} and numerical integration schemes have opened new research avenues for learning extremely accurate dynamical models even from irregularly-sampled training data.

However, for a wide range of real-world systems, we are never provided with an observation operator that forms an embedding of the latent state variables. The key issue is that we do not have any guarantees on the existence of a smooth ODE governing the temporal evolution of our observations in $\mathbb{R}^n$.

From this point of view, the question of finding an appropriate dynamical representation of some observed data may not be this straightforward. The fact that our data may come from some higher-dimensional unobserved governing equation may restrict the above-mentioned state-of-the-art algorithms. The main difficulty is how to go from a low-dimensional observation space to higher-dimensional spaces that provides at least a {one-to-one} mapping between two successive observations. From a geometrical point of view, the time delay theorem \cite{takens_theorem} provides a way to build higher dimensional spaces that preserve the topological properties of the true (unobserved) dynamics limit-cycle from a lower-dimensional phase space. A generalization of this theorem \cite{Sauer1991} shows that we can reconstruct topologically similar limit-cycles using any appropriate smooth composition map of the observations. The derivation of a dynamical system from such representations on the other hand encountered large disparities since no explicit relationships between the defined phase space and an ODE formulation have been clearly made. Classical state-of-the-art techniques such as polynomial representations \cite{brunton_discovering_2016} and K-Nearest Neighbors (KNN) algorithms were proposed but they often fail to achieve both accurate short-term forecasting performance and topologically similar reconstructed limit-cycles.

In this work, we address the identification of a latent embedding that fits an ODE representation of the underlying dynamics of our observations. As detailed in the next Section, the core idea of this work is to replace the augmented observable vector formulated in \cite{takens_theorem} by augmented states that are outputs of an optimization problem with respect to an ODE formulation.


\section{Learning latent representations of partially-observed dynamics}
\label{sec:hrep}


\textbf{Augmented latent dynamics}: Let us assume a continuous $s$-dimensional dynamical system $\mathrm{z}_t$ governed by an autonomous ODE $\Dot{{\mathrm{z}}}_{t} = f({\mathrm{z}}_{t})$ with $\Phi_t$ the corresponding flow $\Phi_t(z_{t_0}) = \int_{t_0}^{t}f(z_{u})du$. In many applications, one cannot fully access to state $\mathrm{z}$ and the observed variables, also called observables, only relate to some components of state $\mathrm{z}$. Formally, we can define an observation function $\mathcal{H}: \mathbb{R}^s \longrightarrow \mathbb{R}^n$ such that observed state $\mathrm{x}_t$ follows $\mathrm{x}_t = \mathcal{H}(\mathrm{z}_t)$

The key question arising here is the extent to which the dynamics expressed in the observation space $\mathbb{R}^n$ reflect the true underlying dynamics in $\mathbb{R}^s$, and consequently, the conditions on $\mathcal{H}$ under which the predictable deterministic dynamical behavior of the hidden states is still predictable in the observation space. To illustrate this issue, we may consider a linear dynamical system in the complex domain governed by the following linear ODE:
\begin{equation}
\left\{
\begin{aligned}
&\Dot{{\mathrm{z}}}_{t} = \alpha{\mathrm{z}}_{t}\\
&{\mathrm{z}}_{t_0} = {\mathrm{z}}_{0}
\label{eq:trig_ODE_exp}
\end{aligned}
\right.
\end{equation}
with ${\mathrm{z}} \in \mathbb{C}$ a state variable and $\alpha \in \mathbb{C}$ a complex imaginary number.
The solution of this problem is 
\begin{align}
&{{\mathrm{z}}}_{t} = Ke^{\alpha t}
\label{eq:sol_trig}
\end{align}
with $K$ a constant depending on ${\mathrm{z}}_{0}$. Assuming that we are only provided the real part as direct measurements ${\mathrm{x}}_t \in \mathbb{R}$ of the true states ${\mathrm{z}}_t$, no smooth autonomous ODE model in the scalar observation space can describe the trajectories of the observed states as the mapping between two successive observed states is not one-to-one. For example, assuming that $\mathrm{z}_{t_0}$ and $\mathrm{z}_{t_1}$ correspond to two states that have the same real part but distinct imaginary parts, the associated observed states are equal
$\mathrm{x}_{t_0} = \mathrm{x}_{t_1}$. However, the time evolution of the states $\mathrm{z}_{t_0}$ and $\mathrm{z}_{t_1}$ differ if they have different imaginary parts, such that the observed states $\mathrm{x}_{t_0+\delta}$ and $\mathrm{x}_{t_1+\delta}$ after any time increment $\delta$ are no more equal. As a consequence, a given observation may have more than one future state and this behavior can not be represented by a smooth ODE in the observation space.

For a given observation operator $\mathcal{H}$ of a deterministic underlying dynamical system governing $\mathrm{z}_{t}$, Takens's theorem guarantees the existence of an augmented space, defined as a delay embedding of observed state series, in which a one-to-one mapping exists between successive time steps of the observation series \cite{takens_theorem}. Rather than exploring such delay embeddings, we aim to identify an augmented latent space, where latent dynamics are governed by a smooth ODE and can be mapped to observed states. Let us define $\mathrm{X}_t \in \mathbb{R}^{d_E}$ the $d_E$-dimensional augmented latent state with $d_E>n$ as follows:
\begin{align}
{\mathrm{X}_t}^T = [\mathrm{x}_{t}^T, \mathrm{y}_{t}^T]
\label{eq:aug_vect}
\end{align}
with $\mathrm{y}_{t} \in \mathbb{R}^{d_E-n}$ the unobserved component of latent state $\mathrm{X}_t$. We aim to define the augmented latent space according to the following state space model:
\begin{equation}
\left\{
\begin{aligned}
&\dot{{{\mathrm{X}}}}_{t} = f_\theta({\mathrm{X}}_{t})\\
&{{\mathrm{x}}}_{t} = G({\mathrm{X(t)}})
\label{eq:aug_ode}
\end{aligned}
\right.
\end{equation}
where the dynamical operator $f_\theta$ belongs to a family of smooth operators (in order to guarantee uniqueness \cite{coddington1955theory}) parametrized by a parameter vector $\theta$. We typically consider a neutral-network representation with Lipschitz nonlinearities and finite weights. As detailed in the next sections, we address the identification of the operator $f_\theta$ and associated latent space $\mathrm{X}$ from a dataset of observed state series $\{\mathrm{x}_0,\ldots,\mathrm{x}_T\}$ as well as the exploitation of identified latent dynamics for the forecasting of the time evolution of the observed states, for instance unobserved future states  $\{\mathrm{x}_{T+1},\ldots,\mathrm{x}_{T+N}\}$.

\textbf{Learning scheme}: Given an observation time series $\{\mathrm{x}_0,\ldots,\mathrm{x}_T\}$, we aim to identify the state-space model defined by (\ref{eq:aug_ode}), which amounts to learning the parameters $\theta$ of the dynamical operator $f_\theta$. However, as the component $\mathrm{y}_{t}$ of the latent state ${\mathrm{X}_t}$ is never observed, this identification requires the joint optimization of the model parameters $\theta$ as well as the hidden component $\mathrm{y}_{t}$. Formally, this problem is then stated as the following minimization of the forecasting error on observed variables:
\begin{equation}
\begin{array}{c}
\hat{\theta} = \arg \displaystyle \min_{\theta}
    \displaystyle \min_{\{y_t\}_{t}} \sum_{t=1}^T \| \mathrm{x}_{t} - G\left (\Phi_{\theta,t} \left(X_{t-1}) \right) \right ) \|^2 \\~\\
    \mbox{Subject to  } 
    \left \{
\begin{array}{lcl}
    X_t& =& \Phi_{\theta,t}(X_{t-1})\\~\\
    G(X_{t})& =& \mathrm{x}_{t}\\
    \end{array}\right.
    \end{array}
\label{eq:opti_crit}
\end{equation}
with  $\Phi_{\theta,t}$ the one-step-ahead diffeomorphic mapping associated with operator $f_\theta$ such that: 
$$\Phi_{\theta,t}(X_{t-1}) = X_{t-1} + \int_{t-1}^{t}f_\theta(X_{u})du$$
In (\ref{eq:opti_crit}), the loss to be minimized involves the one-step-ahead forecasting error for the observed variable $\mathrm{x}_{t}$. The constraints state that the augmented state $\{X_t\}_{t}$ is composed of observed component $G(X_{t})$ and 
should be a solution of ODE (\ref{eq:aug_ode}). Here, we numerically minimize the equivalent formulation:
\begin{equation}
 \displaystyle \min_{\theta}
    \displaystyle \min_{\{\mathrm{y}_{t}\}_{t}} \sum_{t=1}^T \| \mathrm{x}_{t} - G\left (\Phi_{\theta,t} \left(X_{t-1}) \right) \right ) \|^2 
    + \lambda \| X_t -  \Phi_{\theta,t}(X_{t-1})\|^2 \\~\\
\label{eq:opti_crit 2}
\end{equation}
where $X_{t}^T=[\mathrm{x}_{t}^T,\mathrm{y}_{t}^T]$ and $\lambda$ a weighting parameter. The term $ \| X_t -  \Phi_{\theta,t}(X_{t-1})\|^2$ may be regarded as a regularization term such that the inference of the unobserved component $\mathrm{y}_{t}$ of the augmented state $X_{t-1}$ is not solved independently for each time step.

Using a neural-network parametrization for the ODE operator $f_\theta$, the corresponding forecasting operator $\Phi_{\theta,t}$ is also stated as a neural network based on a numerical integration scheme formulation (typically a $4^{th}$-order Runge-Kutta scheme). This architecture, very much similar to a ResNet \cite{he_deep_2015}, allows very accurate identification of ODE models \cite{fablet_blin_ieee,ResINN}. Hence, for a given observed state series $\{x_0,\ldots,x_T\}$, we minimize ($\ref{eq:opti_crit 2}$) jointly w.r.t. $\theta$ and unobserved variables $\{\mathrm{y}_0,\ldots,\mathrm{y}_T\}$. In the experiments reported in Section \ref{sec:exp}, we consider bilinear architectures \cite{fablet_blin_ieee}. The proposed framework applies to any neural-network architecture.

\textbf{Application to forecasting}: We also apply the proposed framework to the forecasting of the observed states $\mathrm{x}_t$. Given a trained latent dynamical model (\ref{eq:aug_ode}), forecasting future states for  $\mathrm{x}_t$ relies on the forecasting of the entire latent state $X_t$. The latter amounts to determining an initial condition of the unobserved component $\mathrm{y}_t$ and performing a numerical integration of the trained ODE (\ref{eq:aug_ode}). 

Let us denote by $\mathrm{x}^n_{t},$ $t \in \left\{t_0,...,T\right\}$ a new series of observed states. We aim to forecast future states $\mathrm{x}^n_{t},$ $t \in \left\{T+1,...,T+\delta T\right\}$. Following (\ref{eq:opti_crit 2}), we infer the unobserved component $\widehat{\mathrm{y}}_{T}$ of latent state $X^n_{T}$ at time $T$ from the minimization of: 
\begin{equation}
\hat{\mathrm{y}}^n_{T} = \arg \min_{\mathrm{y}^n_{T}}  \displaystyle \min_{\{\mathrm{y}^n_{t}\}_{t<T}} \sum_{t=T+1}^{T+\delta T} \| \mathrm{x}^n_{t} - G\left (\Phi_{\theta,t} \left(X^n_{t-1}) \right) \right ) \|^2 
    + \lambda \| X^n_t -  \Phi_{\theta,t}(X^n_{t-1})\|^2
\label{eq:opti_crit 3}
\end{equation}
Here, we only minimize w.r.t. latent variables $\{\mathrm{y}^n_{t}\}$ given 
the trained forecasting operator $\Phi_{\theta,t}$. This minimization relates to a variational assimilation issue with partially-observed states and known dynamical and observation operators \cite{Lynch2010}. Similarly to the learning step, we benefit from the neural-network parameterization of operator $\Phi_{\theta,t}$ and from the associated automatic differentiation tool to compute the solution of the above minimization using a gradient descent. 

We may consider different initialization strategies for this minimization problem. 
Besides a simple random initialization, we may benefit from the information gained  on the manifold spanned by the unobserved components during the training stage. The basic idea comes to assume that the training dataset is likely to comprise state trajectories which are similar to the new one. As the training step embeds the inference of the whole latent state sequence, we may pick as initialization for minimization (\ref{eq:opti_crit 3}) the inferred augmented latent state in the training dataset which leads to the observed state trajectory that is the most similar (in the sense of the L2 norm) to the new observed sequence $\mathrm{x}^n_{t},$. The interest of this initialization scheme is two-fold: (i) speeding-up the convergence of minimization (\ref{eq:opti_crit 3}) as we expect to be closer to the minimum; (ii) considering an initial condition which is  in the basin of attraction of the reconstructed limit-cycle. The latter may be critical as we cannot guarantee that the learnt model does not involve other limit-cycles than the ones truly revealed by the training dataset, which may lead to a convergence to a local and poorly relevant minimum.

\section{Numerical experiments}
\label{sec:exp}



\begin{figure}[h]
\centering
  \includegraphics[clip,width=0.9\columnwidth,height=5cm]{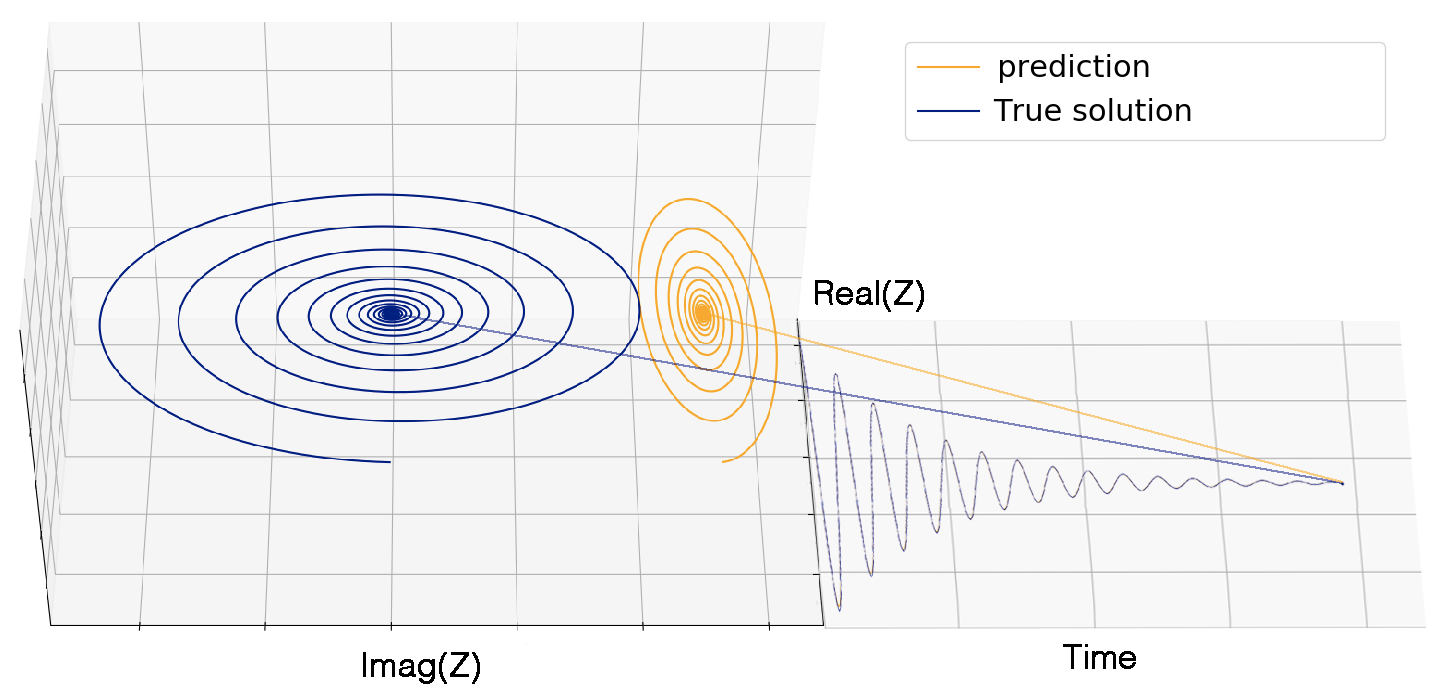}%
\caption{{\bf Inference of latent dynamics for a 2-dimensional linear ODE}: (left) true (blue) and inferred (orange) 2-dimensional latent states, (right) true observation sequence (blue) and predicted observation sequence from the inferred initial condition using the sequence of previous observations (orange). The considered 2-dimensional linear ODE is given by (\ref{eq:trig_ODE}). 
}
\label{fig:forecast_lin}
\end{figure}

\textbf{Application to a linear ODE}: In order to illustrate the key principles of the proposed framework, we consider 
the following linear ODE in the complex domain:
\begin{equation}
\left\{
\begin{aligned}
&\Dot{{\mathrm{z}}}_{t} = \alpha{\mathrm{z}}_{t}\\
&{\mathrm{z}}_{t_0} = {\mathrm{z}}_{0}
\label{eq:trig_ODE}
\end{aligned}
\right.
\end{equation}
with $\alpha = -0.1-0.5j$, $j^2 = -1$ and $\mathrm{z}_{0} = 0.5$. 
As $\alpha \in \mathbb{C}$ with $Real(\alpha)<0$ and $\mathrm{z}_{0} \neq 0$, the solution of this ODE is an ellipse in the complex plane (Fig. \ref{fig:forecast_lin}).

We assume that the observed state is the real part of the underlying state, {\em i.e.} the observation function $\mathcal{H} : \mathbb{C} \longrightarrow \mathbb{R}$ is given by $\mathrm{x}_t = Real(\mathrm{z}_t)$. This is a typical example, where the mapping between two successive observations is not a one-to-one  mapping since all the states that have the same real part lead to the same observation. As explained in section \ref{sec:hrep}, one cannot identify an autonomous ODE model that will reproduce the dynamical behavior of the observations in the observations space.

We apply the proposed framework to this toy example. We consider a 2-dimensional augmented state $\mathrm{X}_t = [\mathrm{x}_{t}, \mathrm{y}^1_{t}]$. As neural-network parametrization for operator $f_\theta$, we consider a neural network with a single linear fully-connected layer. We use an observation series of $10000$ time steps as training data. As illustrated in Fig.\ref{fig:forecast_lin}, the inferred latent state dynamics, though different from the true ones, depicts a similar spiral pattern. 
This result is in agreement with the geometrical reconstruction techniques \cite{takens_theorem} of the latent dynamics up to a diffeomorphic mapping. Overall, our model learns a dynamical behavior similar to the true model represented by an elliptic transient and an equilibrium point limit-set.

We also illustrate the forecasting performance of the proposed framework in Fig. \ref{fig:forecast_lin}. Given the same initial condition over the observable state, we compare the forecasting of the observed state by the true equation and the proposed model. For the proposed model, the initial condition over the unobservable component results from minimization (\ref{eq:opti_crit 3}). The resulting forecast of the observation sequence is similar to the true observation sequence (mean square error < 1E-5).


\textbf{Lorenz-63 dynamics}: Lorenz-63 dynamical system is a 3-dimensional model that involves, under some specific parametrizations \cite{lorenz_deterministic_1963}, chaotic dynamics with a strange attractor. We simulate chaotic Lorenz-63 state sequences with the same model parameters as proposed in \cite{lorenz_deterministic_1963} using the LOSDA ODE solver \cite{odepack} with an integration step of 0.01. We assume that only the first Lorenz-63 variable is observed $\mathrm{x}_t = \mathrm{z}_{t,1}$. We apply the proposed framework to this experimental setting using a training sequence of $10000$ time-steps. 

For benchmarking purposes, we perform a quantitative comparison with state-of-the-art approaches using delay embedding representations \cite{takens_theorem}. The parameters of the delay embedding representation, namely the lag $\tau$ and the dimension $d_E$ of the augmented space were computed using state-of-the-art techniques. Specifically, the lag parameter was computed using both the mutual information and correlation  techniques \cite{takens_params_1}, respectively denoted as $\tau_{MI}$ and  $\tau_{Corr}$. Regarding the dimension of the embedding representation, we used the Takens embedding condition $d_{E} = 2d+1$ with $d$ the dimension of the hidden limit-cycle. The delay embedding dimension was also computed using the False Nearest Neighbors (FNN) method \cite{takens_params_2}. We also tested arbitrary parameters for the delay embedding dimension. Given the delay embedding representation, we tested two state-of-the-art data-driven representations of the dynamics. The Analog Forecasting technique (AF) which is based on the nearest neighbours algorithm \cite{lguensat_analog_2017} and the Sparse Regression (SR) method on a second order polynomial representation of the delay embedding states.
Regarding deep learning models, we evaluate the Latent-ODE model as proposed in \cite{chen2018neural} for the modeling of time series. The architecture of the Latent-ODE model is the following: The recognition network is an RNN with 100 hidden units. The dimension of the latent space is 4 and the dynamical model is a fully-connected network with bilinear layers as proposed in \cite{fablet_blin_ieee} (the same as used in our model). The decoder is a fully connected network with one layer and 15 units.

Regarding the proposed framework, referred to as Neural embedding for Dynamical Systems (NbedDyn), we tested the model for different dimensions of the augmented state space, namely from 3 to 6. The neural-network parametrization for operator $f_{\theta}$ is a fully-connected architecture with bilinear layers as proposed in \cite{fablet_blin_ieee} \footnote{The code will be available on GitHub soon.}.

\begin{table}[h]
\centering
\begin{adjustbox}{width=4.9in,totalheight=2.8in}
\begin{tabular}{lll*{3}c}
\toprule
\multicolumn{3}{c}{Model} &$t_0+h$ &$t_0+4h$  & $\lambda_1$  \\
\midrule \midrule 
\multirow{4}{*}{AF}
& $\tau_{MI}$ =16 & $d_E(FNN) = 3$ &$7.5E-4$  & $2.4E-3$ &$1.16$  \\
& $\tau_{MI}$ =16 & $d_E(Takens) = 6$ &$6.4E-5$  & $1.6E-4$ &$NaN$  \\
& $\tau_{Corr}$ =29 & $d_E(FNN) = 3$ &$0.0011$  & $0.0021$ &$1.19$  \\
& $\tau_{Corr}$ =29 & $d_E(Takens) = 6$ &$0.0025$  & $0.0063$ &$NaN$  \\
& $\tau$ = 6 & $d_E = 3$ &$2.9E-4$  & $5.5E-4$ &$0.95$  \\
& $\tau$ = 10 & $d_E = 3$ &$1.2E-4$  & $0.00104$ &$0.84$  \\
\midrule
\multirow{4}{*}{SR}
& $\tau_{MI}$ =16 & $d_E(FNN) = 3$ &$0.038$  & $0.299$ &$0.084$  \\
& $\tau_{MI}$ =16 & $d_E(Takens) = 6$ &$0.1076$  & $0.2497$ &$NaN$  \\
& $\tau_{Corr}$ =29 & $d_E$(FNN) = 3 &$0.218$  & $1.005$ &$-0.73$  \\
& $\tau_{Corr}$ =29 & $d_E$(Takens) = 6 &$0.1344$  & $0.7841$ &$NaN$  \\
& $\tau$ = 6 & $d_E = 3$ &$0.00185$  & $0.00256$ &$NaN$  \\
& $\tau$ = 10 & $d_E = 3$ &$0.00733$  & $0.06179$ &$0.2535$  \\
\midrule
\multirow{1}{*}{Latent-ODE}
&   &  &$0.0801$  & $0.520$ &$NaN$  \\
\midrule
\multirow{4}{*}{NbedDyn}
&   & $d_E = 3$ &$8.14E-4$  & $2.29E-3$ &$0.82$  \\
&   & $d_E = 4$ &$1.891E-4$  & $6.106E-4$ &$0.960$  \\
&   & $d_E = 5$ &$2.206E-4$  & $1.869E-3$ &$0.822$  \\
&   & $d_E = 6$ &$\textbf{6.820E-6}$  & $\textbf{6.5125E-5}$ &$\textbf{0.87}$  \\
\bottomrule
\end{tabular}
\end{adjustbox}
\caption {{\bf  \em Forecasting performance on the test set of data-driven models for Lorenz-63 dynamics where only the first variable is observed}: first two columns : mean RMSE for different forecasting time steps, third column : largest Lyapunov exponent of a predicted series of length of 10000 time-steps (The true largest Lyapunov exponent of the Lorenz 63 model is 0.91 \cite{Sprott_chaos}).}
\label{tab:for_63}
\vspace{-0.3cm}
\end{table}

\begin{figure}[h]
\includegraphics[clip,width=\columnwidth,height=7cm]{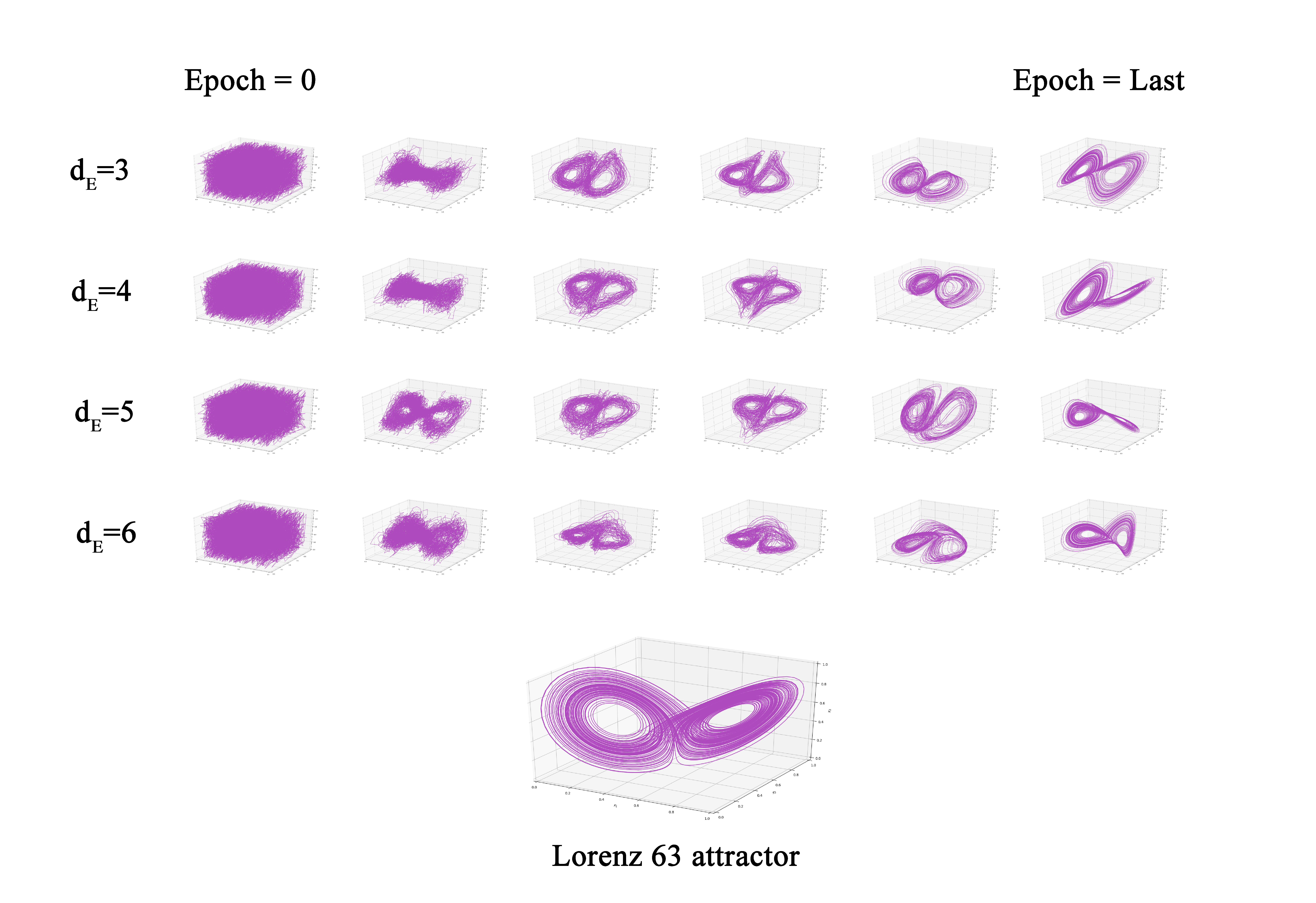}
     \caption{{{\bf \em Evolution of the learnt latent space}: starting from a random initialization of the augmented states $\mathrm{y}_i$, the latent space is optimized according to minimization \ref{eq:opti_crit 2}} to form a  limit-cycle similar to the true Lorenz 63 attractor. We depict 3-dimensional projections of the learnt latent space for the proposed model with different embedding dimensions from $d_E=3$ to $d_E=6$.}
     \label{fig:latent_train}
\end{figure}






Fig. \ref{fig:latent_train} illustrates the learning process for the latent space from the initialization to the last training epoch. We also report the analysis of short-term forecasting performance as well as the long-term asymptotic behavior characterized by the largest Lyapunov exponent of the benchmarked models in Tab \ref{tab:for_63}. The proposed model leads to significant improvements in terms of short term forecasting performance with respect to the other approaches. Surprisingly, The Latent-ODE model lead to the poorest performance both in terms of forecasting error and asymptotic behavior. This is mainly due to the fact that the latent space is seen as a non linear projection of the observed variables through the optimization of the ELBO loss \cite{krishnan_structured_2016}. By contrast, our latent embedding formulation optimizes the latent states to forecast the observed variables which explicitly constrain the latent space to be an embedding of the true underlying dynamics. The SR model seems to lead to better short term forecast, however, it does not capture well the chaotic patterns, which are associated to a positive largest Lyapunov exponent. This may suggest the combination of the SR model and a delay embedding may require additional investigation as a good geometrical reconstruction of the phase space as stated in Takens' theorem does not guarantee the existence of a parametric ODE model based on the corresponding delay embedding variables. Better performance is reported using an analog forecasting approach. The performance however greatly varies depending on the considered definition of the delay embedding. Using ad hoc parameters ($\tau_{MI}$ = 6, $d_E = 3)$, one may retrieve the expected long-term chaotic behavior ($\lambda_1=0.84$) with a relatively low short-term forecasting error (1.2e-4 for a one-step-ahead forecast). When considering the proposed model, we report for all parametrizations, augmented space dimensions from 3 to 6, performance at least in the same range as the best analog forecasting setting. Besides, when increasing the dimension of the augmented space, we significantly decrease short-term forecasting errors (<1.e-5 for a one-step-ahead forecast when considering $d_E=6$, i.e. one order of magnitude compared to the second best model) while keeping an appropriate chaotic long-term pattern ($\lambda_1=0.87$). 

\textbf{Modeling Sea Level Anomaly (SLA)}: The data driven identification of dynamical representations of real data is an extremely difficult task especially when the underlying processes involve non stable behaviors such as chaotic attractors. This is mainly due to the fact that we  do not have any exact knowledge of the closed form of the equations governing the temporal evolution of our variables. Furthermore, the measured quantity may depend on other unobserved variables which makes the exploitation of data-driven techniques highly challenging.

In this context, we report an application to SLA (Sea Level Anomaly) dynamics, which relate to upper ocean dynamics and are monitored by satellite altimeters \cite{Calmant2008}. Sea surface dynamics are chaotic dynamics and clearly involve latent processes, typically subsurface and atmospheric processes. The dataset used in our experiments is a SLA time series obtained using the WMOP product \cite{WMOP_soc}. The spatial resolution of our data is a $0.05\degree$ and the temporal resolution $h=1$ day. We use the data from January 2009 to December 2014 as training data and we tested our approach on the last month of the year 2014. The considered region is located on south Mallorca ($2.5\degree E$ to $4.25\degree E$, $37.25\degree N$ to $39.5\degree N$). The spatial scales of the data were decomposed through a PCA decomposition of dimension $N_E=15$, which amounts to capture $92\%$ of the total variance.

\begin{wraptable}{r}{0.5\linewidth}
\centering
\begin{adjustbox}{width=2.6in,totalheight=1.1in}
\begin{tabular}{ll*{6}c}
\toprule
\multicolumn{1}{c}{Model} & &$t_0+h$ & $t_0+2h$ & $t_0+4h$ \\
\midrule \midrule 
\multirow{2}{*}{Latent-ODE}
 &RMSE & 0.025 & 0.032 & 0.048 \\
 &Corr & 89.25\% & 89.85\% & 89.07\% \\
\midrule
\multirow{2}{*}{AF}
&RMSE& 0.046  & 0.062 & 0.073 \\  
&Corr& 95.56\%  & 86.11\% & 69.87\% \\  
\midrule
\multirow{2}{*}{NbedDyn}
& RMSE &  0.002 & 0.007 & 0.027 \\  
& Corr &  99.99 \% & 99.90\% & 98.95\% \\  

\bottomrule
\end{tabular}
\end{adjustbox}
\caption {{\bf  \em SLA Forecasting performance on the test set of data-driven models}: RMSE and correlation coefficients for different forecasting time steps.}
\label{tab:for_SLA}
\end{wraptable}

We report forecasting performance for our model and include a comparison with analog methods and a neural ODE setting in Tab. \ref{tab:for_SLA}. For our model, we consider the following parameterization: a latent space with $d_E = 60$ and a fully connected bilinear architecture with 60 bilinear neurons and 4~fully-connected layers for the dynamical model $f_\theta$. Our model clearly outperforms the three benchmarked schemes with a very significant gain for the forecasting performance at one day (relative gain greater than 90 \%) and two days (relative gain greater than 90 \%). For a 4-day-ahead forecasting, our model still outperforms the other ones though the gain is lower (relative gain of 40\%). We let the reader refer to the Supplementary Material for a more detailed analysis of these experiments, including visual comparisons of the forecasts.


\section{Discussion}
\label{sec:disc}

In this work, we address the data-driven identification of latent dynamics for systems which are only partially observed, {\em i.e.} when some components of the system of interest are never observed. The reported forecasting performance for Lorenz-63 dynamics is in line with the forecasting performance of state-of-the-art learning-based approaches for a noise-free and fully-observed setting. This is of key interest for real-world applications, where observing systems most often monitor only some components of the underlying systems. As a typical example, the SLA forecasting experiment clearly motivates the proposed framework in the context of ocean dynamics for which neither in situ nor satellite observing systems can provide direct observations for all state variables (e.g., subsurface velocities, fine-scale sea surface currents). 

We may also further discuss how the proposed framework relates to state-of-the-art dynamical system theory approaches. Most of these approaches rely on delay embedding, as Takens's theorem states the existence of a delay embedding in which the topological properties of the hidden dynamical system are equivalent to those of the true systems up to a diffeomorphic mapping. Hence, state-of-the-art approaches typically combine the selection of a delay embedding representation within classic regression models to represent the one-step-ahead mapping in the considered embedding. Here, we consider latent dynamics governed by an unknown ODE (\ref{eq:aug_ode}) but we do not explicitly state the latent space. This is however implicit in our forecasting framework. By construction, the considered forecasting model relies on the integration of the learnt ODE (\ref{eq:aug_ode}) from an initial condition given as the solution of minimization (\ref{eq:opti_crit 3}). Let us consider the following embedding $\psi$ such that:
\begin{equation}
    \psi \left( \{\mathrm{x}_{t}\}_{t_0:T}\right) = 
    \arg \min_{\mathrm{X}_{T}}  \displaystyle \min_{\{\mathrm{X}_{t}\}_{t<T}} \sum_{t=1}^T \| \mathrm{x}_{t} - G\left (\Phi_{\theta,t} \left(X_{t-1}) \right) \right ) \|^2 
    + \lambda \| X_t -  \Phi_{\theta,t}(X_{t-1})\|^2
\end{equation}
Given this embedding, the resulting one-step-ahead forecasting for the observed variable may written as:
\begin{equation}
\mathrm{x}_{T+1} = G\left ( \Phi_{\theta,t} \left( \psi \left( \{\mathrm{x}_{t}\}_{t=t_0:T}\right) \right ) \right )
\end{equation}
Hence, $\psi$ defines a delay embedding representation implicitly stated through minimization (\ref{eq:opti_crit 3}). In this embedding, the dynamics of the observed system $\mathrm{x}$ is governed by the composition of observation operator $G$ and forecasting operator $\Phi_{\theta,t}$. Regarding the literature on Koopman operator theory, most approaches rely on the explicit identification of eigenfunctions and eigenvalues of the Koopman operator \cite{KoopmanGeneral,brunton2016koopman,DMD_koopman}. Our framework relates to the identification of the infinitesimal generator $f_\theta$ of the one-parameter subgroup defined by Koopman operator through the ODE representation (\ref{eq:aug_ode}). By construction, the Koopman operator associated with the identified operator $f_{\hat{\theta}}$ is also diagonalizable, such that the identification of infinitesimal generator $f_{\hat{\theta}}$ provides an implicit decomposition of the Koopman operator of the underlying and unknown dynamical system onto the eigenbasis of the learnt latent dynamics governed by ODE (\ref{eq:aug_ode}).

Future work will further explore methodological aspects, especially the application to high-dimensional and stochastic systems. In the considered framework, operator $G$ is stated as an identity operator on the observed component of state $X_t$. For high-dimensional systems, operator $G$ may be stated as an encoder to keep the latent space low-dimensional and make minimization (\ref{eq:opti_crit 2}) tractable. The combination of the proposed framework with the variational setting considered in  the Latent-ODE model \cite{chen2018neural} also appears as an interesting direction for future work.  
The extension to stochastic systems through the identification of a Stochastic ODE is also of key interest, for instance for future applications of the proposed framework to geophysical random flows, especially to the simulation and forecasting of ocean-atmosphere dynamics in which stochastic components naturally arise \cite{chapron2018large}.  

\bibliographystyle{IEEEtran}
\bibliography{biblio}

\begin{thebibliography}{10}
\providecommand{\url}[1]{#1}
\csname url@samestyle\endcsname
\providecommand{\newblock}{\relax}
\providecommand{\bibinfo}[2]{#2}
\providecommand{\BIBentrySTDinterwordspacing}{\spaceskip=0pt\relax}
\providecommand{\BIBentryALTinterwordstretchfactor}{4}
\providecommand{\BIBentryALTinterwordspacing}{\spaceskip=\fontdimen2\font plus
\BIBentryALTinterwordstretchfactor\fontdimen3\font minus
  \fontdimen4\font\relax}
\providecommand{\BIBforeignlanguage}[2]{{%
\expandafter\ifx\csname l@#1\endcsname\relax
\typeout{** WARNING: IEEEtran.bst: No hyphenation pattern has been}%
\typeout{** loaded for the language `#1'. Using the pattern for}%
\typeout{** the default language instead.}%
\else
\language=\csname l@#1\endcsname
\fi
#2}}
\providecommand{\BIBdecl}{\relax}
\BIBdecl

\bibitem{ident_control}
\BIBentryALTinterwordspacing
T.~L. Lai and C.~Z. Wei, ``Least squares estimates in stochastic regression
  models with applications to identification and control of dynamic systems,''
  \emph{The Annals of Statistics}, vol.~10, no.~1, pp. 154--166, 1982.
  [Online]. Available: \url{http://www.jstor.org/stable/2240506}
\BIBentrySTDinterwordspacing

\bibitem{Abarbanel1996}
\BIBentryALTinterwordspacing
H.~D.~I. Abarbanel and U.~Lall, ``Nonlinear dynamics of the great salt lake:
  system identification and prediction,'' \emph{Climate Dynamics}, vol.~12,
  no.~4, pp. 287--297, Mar 1996. [Online]. Available:
  \url{https://doi.org/10.1007/BF00219502}
\BIBentrySTDinterwordspacing

\bibitem{jeong_hussain_1995}
J.~Jeong and F.~Hussain, ``On the identification of a vortex,'' \emph{Journal
  of Fluid Mechanics}, vol. 285, p. 69–94, 1995.

\bibitem{iden_economics}
\BIBentryALTinterwordspacing
T.~C. Koopmans, ``Identification problems in economic model construction,''
  \emph{Econometrica}, vol.~17, no.~2, pp. 125--144, 1949. [Online]. Available:
  \url{http://www.jstor.org/stable/1905689}
\BIBentrySTDinterwordspacing

\bibitem{brunton_discovering_2016}
\BIBentryALTinterwordspacing
S.~L. Brunton, J.~L. Proctor, and J.~N. Kutz,
  ``\BIBforeignlanguage{en}{Discovering governing equations from data by sparse
  identification of nonlinear dynamical systems},''
  \emph{\BIBforeignlanguage{en}{Proceedings of the National Academy of
  Sciences}}, vol. 113, no.~15, pp. 3932--3937, Apr. 2016. [Online]. Available:
  \url{http://www.pnas.org/lookup/doi/10.1073/pnas.1517384113}
\BIBentrySTDinterwordspacing

\bibitem{fablet_blin_ieee}
R.~{Fablet}, S.~{Ouala}, and C.~{Herzet}, ``Bilinear residual neural network
  for the identification and forecasting of geophysical dynamics,'' in
  \emph{2018 26th European Signal Processing Conference (EUSIPCO)}, Sep. 2018,
  pp. 1477--1481.

\bibitem{nguyen2019emlike}
\BIBentryALTinterwordspacing
D.~Nguyen, S.~Ouala, L.~Drumetz, and R.~Fablet, ``Em-like learning chaotic
  dynamics from noisy and partial observations,'' \emph{SciRate}, Mar. 2019.
  [Online]. Available: \url{https://scirate.com/arxiv/1903.10335}
\BIBentrySTDinterwordspacing

\bibitem{takens_theorem}
F.~Takens, ``Detecting strange attractors in turbulence,'' in \emph{Dynamical
  Systems and Turbulence, Warwick 1980}, D.~Rand and L.-S. Young, Eds.\hskip
  1em plus 0.5em minus 0.4em\relax Berlin, Heidelberg: Springer Berlin
  Heidelberg, 1981, pp. 366--381.

\bibitem{Abarbanel_model_chaos}
\BIBentryALTinterwordspacing
H.~D.~I. Abarbanel, \emph{Modeling Chaos}.\hskip 1em plus 0.5em minus
  0.4em\relax New York, NY: Springer New York, 1996, pp. 95--114. [Online].
  Available: \url{https://doi.org/10.1007/978-1-4612-0763-4_6}
\BIBentrySTDinterwordspacing

\bibitem{paduart_identification_2010}
\BIBentryALTinterwordspacing
J.~Paduart, L.~Lauwers, J.~Swevers, K.~Smolders, J.~Schoukens, and R.~Pintelon,
  ``Identification of nonlinear systems using {Polynomial} {Nonlinear} {State}
  {Space} models,'' \emph{Automatica}, vol.~46, no.~4, pp. 647--656, Apr. 2010.
  [Online]. Available:
  \url{http://www.sciencedirect.com/science/article/pii/S000510981000021X}
\BIBentrySTDinterwordspacing

\bibitem{NN_takens}
\BIBentryALTinterwordspacing
J.~Frank, S.~Mannor, and D.~Precup, ``Activity and gait recognition with
  time-delay embeddings,'' in \emph{Proceedings of the Twenty-Fourth AAAI
  Conference on Artificial Intelligence}, ser. AAAI'10.\hskip 1em plus 0.5em
  minus 0.4em\relax AAAI Press, 2010, pp. 1581--1586. [Online]. Available:
  \url{http://dl.acm.org/citation.cfm?id=2898607.2898859}
\BIBentrySTDinterwordspacing

\bibitem{SVM_takens}
\BIBentryALTinterwordspacing
A.~Kazem, E.~Sharifi, F.~K. Hussain, M.~Saberi, and O.~K. Hussain, ``Support
  vector regression with chaos-based firefly algorithm for stock market price
  forecasting,'' \emph{Applied Soft Computing}, vol.~13, no.~2, pp. 947 -- 958,
  2013. [Online]. Available:
  \url{http://www.sciencedirect.com/science/article/pii/S1568494612004449}
\BIBentrySTDinterwordspacing

\bibitem{takens_params_1}
\BIBentryALTinterwordspacing
H.~D.~I. Abarbanel, \emph{Choosing Time Delays}.\hskip 1em plus 0.5em minus
  0.4em\relax New York, NY: Springer New York, 1996, pp. 25--37. [Online].
  Available: \url{https://doi.org/10.1007/978-1-4612-0763-4_3}
\BIBentrySTDinterwordspacing

\bibitem{takens_params_2}
\BIBentryALTinterwordspacing
------, \emph{Choosing the Dimension of Reconstructed Phase Space}.\hskip 1em
  plus 0.5em minus 0.4em\relax New York, NY: Springer New York, 1996, pp.
  39--67. [Online]. Available:
  \url{https://doi.org/10.1007/978-1-4612-0763-4_4}
\BIBentrySTDinterwordspacing

\bibitem{KoopmanGeneral}
\BIBentryALTinterwordspacing
B.~O. Koopman, ``Hamiltonian systems and transformations in hilbert space,''
  \emph{Proceedings of the National Academy of Sciences of the United States of
  America}, vol.~17, no.~5, pp. 315--318, 1931. [Online]. Available:
  \url{http://www.jstor.org/stable/86114}
\BIBentrySTDinterwordspacing

\bibitem{Sauer1991}
\BIBentryALTinterwordspacing
T.~Sauer, J.~A. Yorke, and M.~Casdagli, ``Embedology,'' \emph{Journal of
  Statistical Physics}, vol.~65, no.~3, pp. 579--616, Nov 1991. [Online].
  Available: \url{https://doi.org/10.1007/BF01053745}
\BIBentrySTDinterwordspacing

\bibitem{wiewel2018latentspace}
S.~Wiewel, M.~Becher, and N.~Thuerey, ``Latent-space physics: Towards learning
  the temporal evolution of fluid flow,'' 2018.

\bibitem{raissi2018multistep}
M.~Raissi, P.~Perdikaris, and G.~E. Karniadakis, ``Multistep neural networks
  for data-driven discovery of nonlinear dynamical systems,'' \emph{arXiv
  preprint arXiv:1801.01236}, 2018.

\bibitem{chen2018neural}
T.~Q. Chen, Y.~Rubanova, J.~Bettencourt, and D.~K. Duvenaud, ``Neural ordinary
  differential equations,'' in \emph{Advances in Neural Information Processing
  Systems}, 2018, pp. 6571--6583.

\bibitem{ResINN}
S.~{Ouala}, A.~{Pascual}, and R.~{Fablet}, ``Residual integration neural
  network,'' in \emph{ICASSP 2019 - 2019 IEEE International Conference on
  Acoustics, Speech and Signal Processing (ICASSP)}, May 2019, pp. 3622--3626.

\bibitem{coddington1955theory}
E.~A. Coddington and N.~Levinson, \emph{Theory of ordinary differential
  equations}.\hskip 1em plus 0.5em minus 0.4em\relax Tata McGraw-Hill
  Education, 1955.

\bibitem{he_deep_2015}
\BIBentryALTinterwordspacing
K.~He, X.~Zhang, S.~Ren, and J.~Sun, ``Deep {Residual} {Learning} for {Image}
  {Recognition},'' \emph{arXiv:1512.03385 [cs]}, december 2015, arXiv:
  1512.03385. [Online]. Available: \url{http://arxiv.org/abs/1512.03385}
\BIBentrySTDinterwordspacing

\bibitem{Lynch2010}
\BIBentryALTinterwordspacing
P.~Lynch and X.-Y. Huang, \emph{Initialization}.\hskip 1em plus 0.5em minus
  0.4em\relax Berlin, Heidelberg: Springer Berlin Heidelberg, 2010, pp.
  241--260. [Online]. Available:
  \url{https://doi.org/10.1007/978-3-540-74703-1_9}
\BIBentrySTDinterwordspacing

\bibitem{lorenz_deterministic_1963}
\BIBentryALTinterwordspacing
E.~N. Lorenz, ``Deterministic {Nonperiodic} {Flow},'' \emph{Journal of the
  Atmospheric Sciences}, vol.~20, no.~2, pp. 130--141, Mar. 1963. [Online].
  Available:
  \url{http://journals.ametsoc.org/doi/abs/10.1175/1520-0469(1963)020%3C0130:DNF%3E2.0.CO;2}
\BIBentrySTDinterwordspacing

\bibitem{odepack}
A.~C. Hindmarsh, ``{ODEPACK}, a systematized collection of {ODE} solvers,''
  \emph{IMACS Transactions on Scientific Computation}, vol.~1, pp. 55--64,
  1983.

\bibitem{lguensat_analog_2017}
\BIBentryALTinterwordspacing
R.~Lguensat, P.~Tandeo, P.~Ailliot, M.~Pulido, and R.~Fablet,
  ``\BIBforeignlanguage{en}{The {Analog} {Data} {Assimilation}},''
  \emph{\BIBforeignlanguage{en}{Monthly Weather Review}}, aug 2017. [Online].
  Available: \url{http://journals.ametsoc.org/doi/10.1175/MWR-D-16-0441.1}
\BIBentrySTDinterwordspacing

\bibitem{Sprott_chaos}
J.~C. Sprott, \emph{Chaos and Time-Series Analysis}.\hskip 1em plus 0.5em minus
  0.4em\relax New York, NY, USA: Oxford University Press, Inc., 2003.

\bibitem{krishnan_structured_2016}
\BIBentryALTinterwordspacing
R.~G. Krishnan, U.~Shalit, and D.~Sontag, ``Structured {Inference} {Networks}
  for {Nonlinear} {State} {Space} {Models},'' \emph{arXiv:1609.09869 [cs,
  stat]}, Sep. 2016, arXiv: 1609.09869. [Online]. Available:
  \url{http://arxiv.org/abs/1609.09869}
\BIBentrySTDinterwordspacing

\bibitem{Calmant2008}
\BIBentryALTinterwordspacing
S.~Calmant, F.~Seyler, and J.~F. Cretaux, ``Monitoring continental surface
  waters by satellite altimetry,'' \emph{Surveys in Geophysics}, vol.~29,
  no.~4, pp. 247--269, Oct 2008. [Online]. Available:
  \url{https://doi.org/10.1007/s10712-008-9051-1}
\BIBentrySTDinterwordspacing

\bibitem{WMOP_soc}
\BIBentryALTinterwordspacing
M.~Juza, B.~Mourre, L.~Renault, S.~Gómara, K.~Sebastián, S.~Lora, J.~P.
  Beltran, B.~Frontera, B.~Garau, C.~Troupin, M.~Torner, E.~Heslop, B.~Casas,
  R.~Escudier, G.~Vizoso, and J.~Tintoré, ``Socib operational ocean
  forecasting system and multi-platform validation in the western mediterranean
  sea,'' \emph{Journal of Operational Oceanography}, vol.~9, no. sup1, pp.
  s155--s166, 2016. [Online]. Available:
  \url{https://doi.org/10.1080/1755876X.2015.1117764}
\BIBentrySTDinterwordspacing

\bibitem{brunton2016koopman}
S.~L. Brunton, B.~W. Brunton, J.~L. Proctor, and J.~N. Kutz, ``Koopman
  invariant subspaces and finite linear representations of nonlinear dynamical
  systems for control,'' \emph{PloS one}, vol.~11, no.~2, p. e0150171, 2016.

\bibitem{DMD_koopman}
J.~H. Tu, C.~W. Rowley, D.~M. Luchtenburg, S.~L. Brunton, and J.~N. Kutz, ``On
  dynamic mode decomposition: Theory and applications,'' \emph{Journal of
  Computational Dynamics}, vol.~1, no.~2, pp. 391--421, 2014.

\bibitem{chapron2018large}
B.~Chapron, P.~D{\'e}rian, E.~M{\'e}min, and V.~Resseguier, ``Large-scale flows
  under location uncertainty: a consistent stochastic framework,''
  \emph{Quarterly Journal of the Royal Meteorological Society}, vol. 144, no.
  710, pp. 251--260, 2018.

\end{thebibliography}
\newpage
\section*{Appendix}
\appendix
\section{Dimensionality analysis of the NbedDyn model}
\label{sec:dim_an}
One of the Key parameters of the proposed model is the dimension of the latent space. Despite the fact that it is extremely challenging to get a prior idea of the dimension of the model in the case of real data experiments, one can analyze the spawned manifold of the learnt latent states to get an idea of the true dimension of the underlying model (true here stands for a sufficient dimension of the latent space). The idea here is to compute the modulus of the eigenvalues of the Jacobian matrix for each input of the training data. An eigenvalue does not influence the temporal evolution of the latent state if it has a modulus that tend to zero. The number of non-zero eigenvalues can then be seen as a sufficient dimension of the latent space.

\begin{figure}[h]
\centering
\subfloat[Eigenvalues real part.]
{
    \includegraphics[scale=.12]{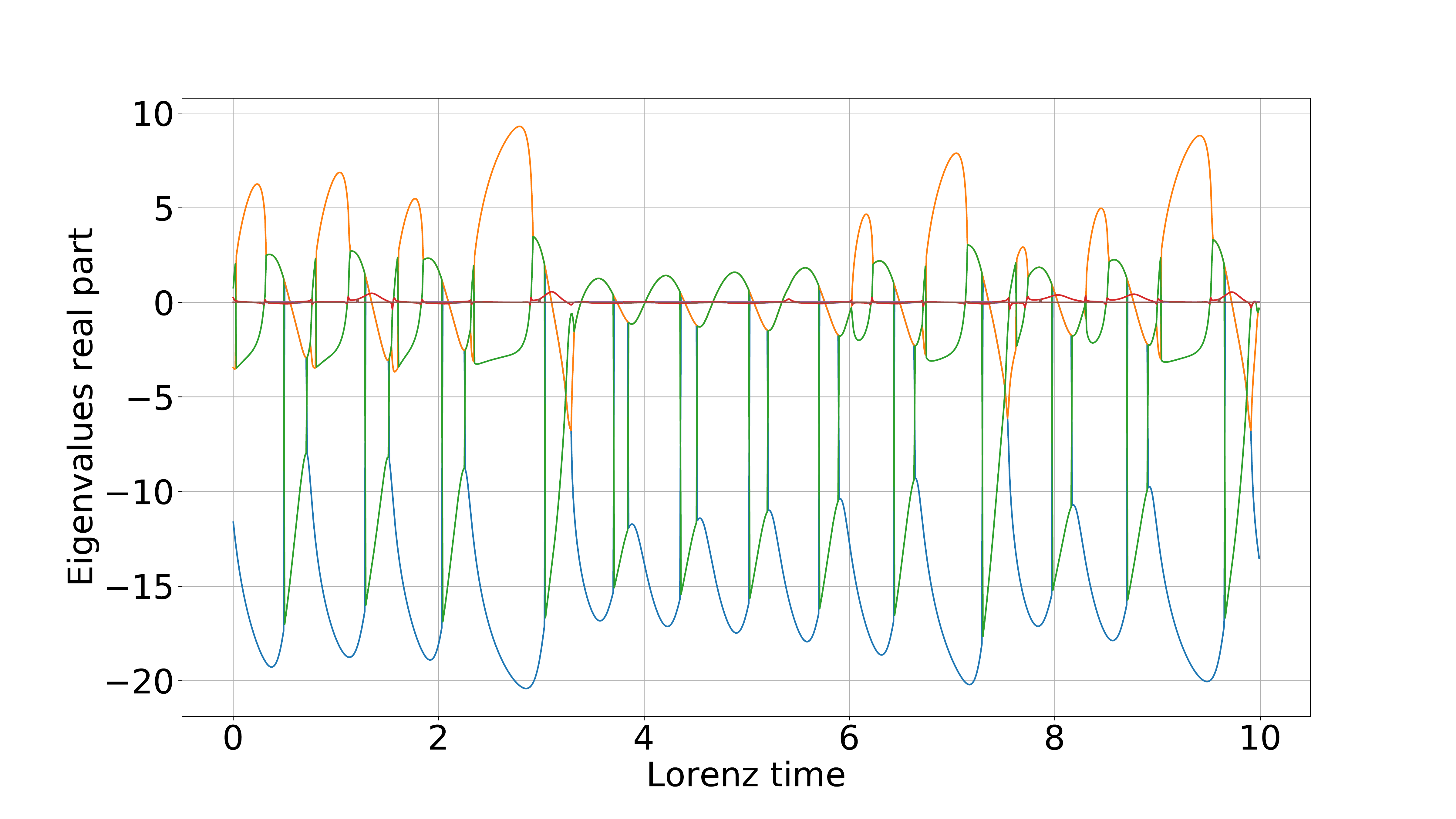}
    \label{fig:real_lor}
}
\subfloat[Eigenvalues imaginary part.]
{
    \includegraphics[scale=.12]{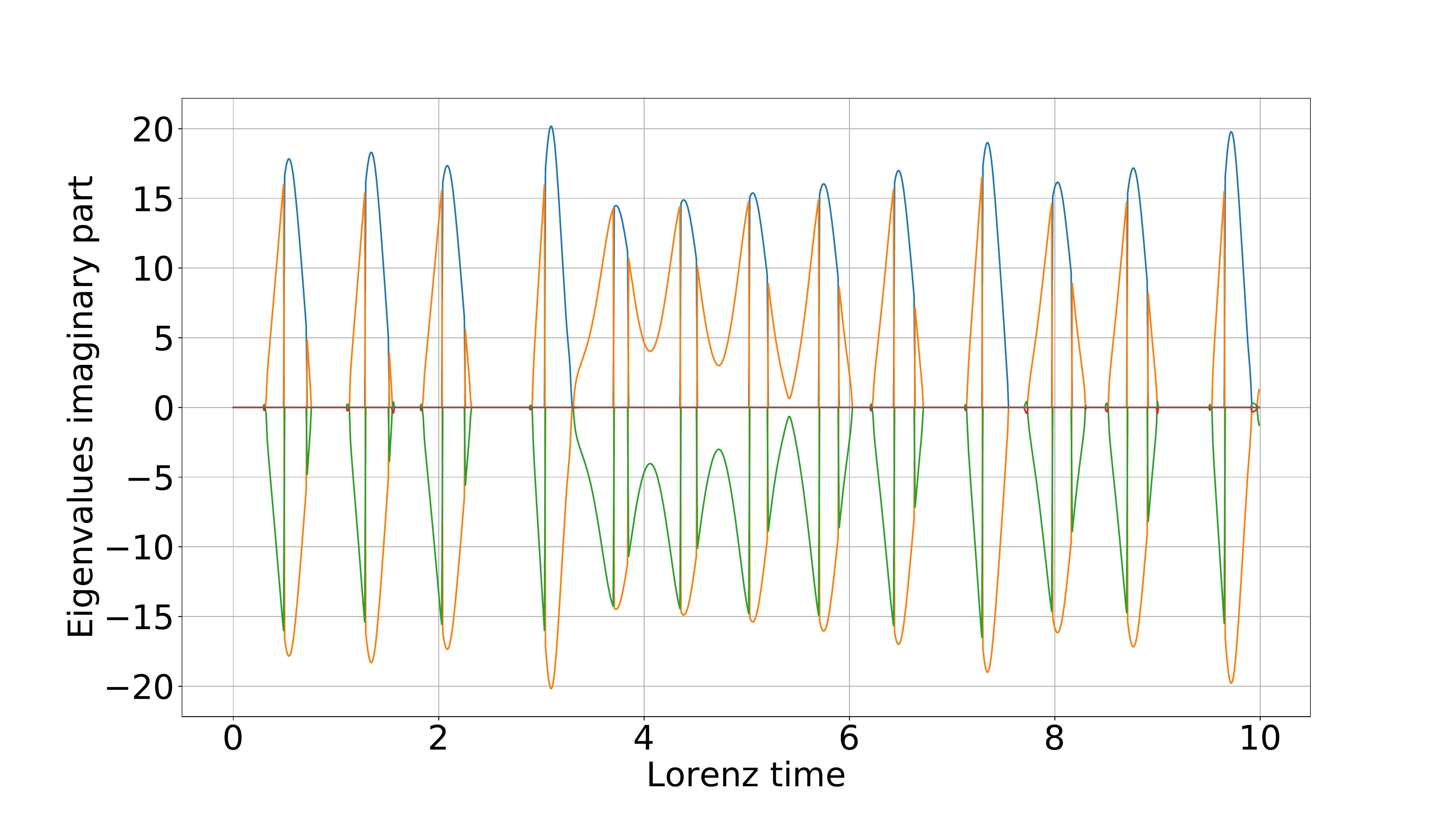}
    \label{fig:imag_lor}
}

\subfloat[Eigenvalues modulus.]{%
  \includegraphics[width=8cm,height=5cm]{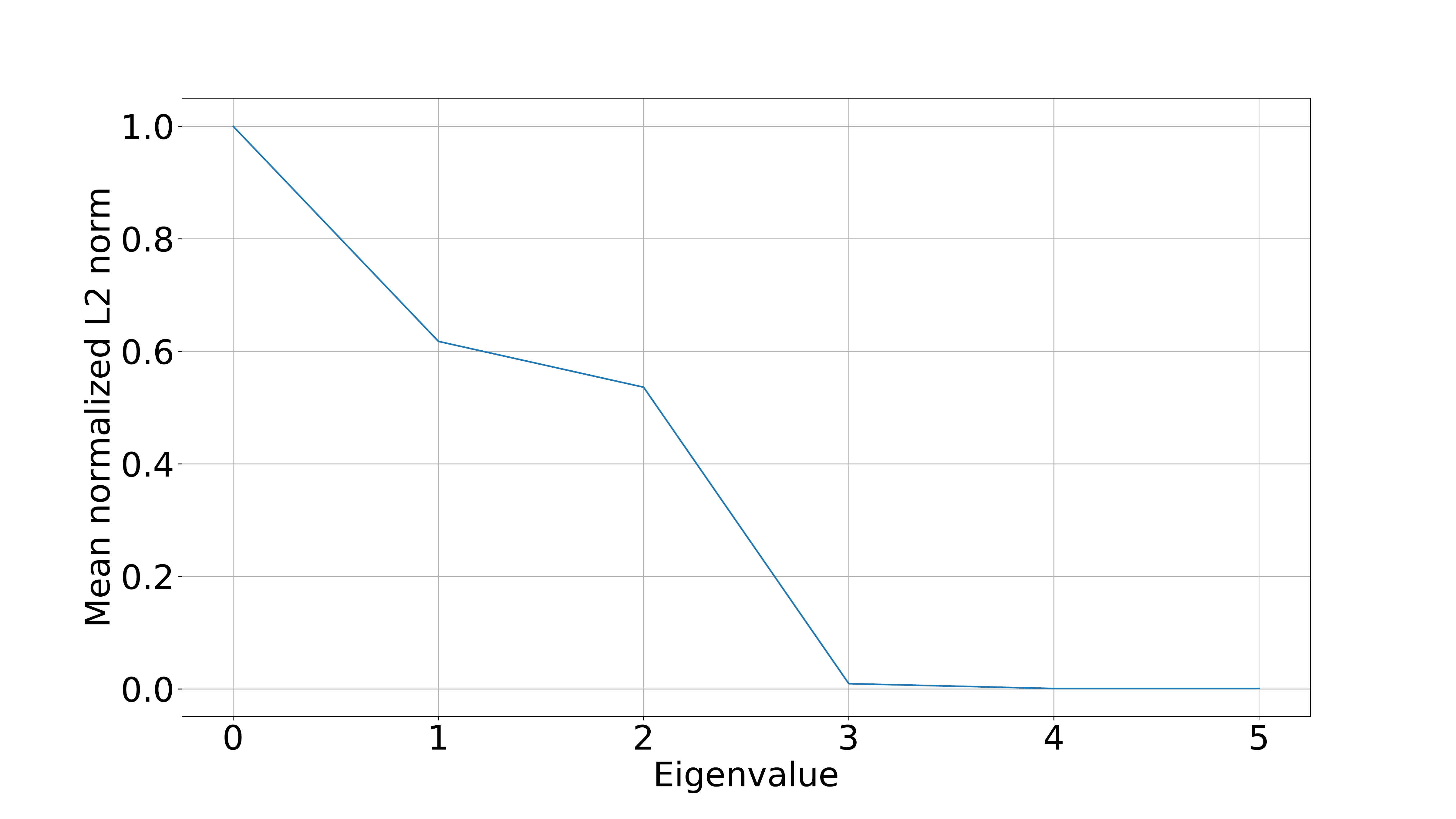}%
  \label{fig:mod_lor}
}

\caption{{{\bf   \em Analysis of the eigenvalues of the NbedDyn model Jacobian matrix.}: Lorenz-63 case-study with $d_E = 6$}}
\label{fig:lorenz_eig}
\end{figure}

Regarding the identification of an ODE model governing the first state variable of the Lorenz 63 model, Fig. \ref{fig:lorenz_eig} illustrates the eigenvalues of the Jacobian matrix and their modulus for a dimension of the latent space $d_E = 6$. Interestingly, only 3 eigenvalues have non-zero modulus and are effectively influencing the underlying dynamics. This result shows that one can use a 3 dimensional latent-space as a sufficient dimension to identify an ODE model governing the first state of the Lorenz 63 system which is the same dimension as the true Lorenz 63 model.

\begin{figure}[h]
\centering
\subfloat[Eigenvalues real part.]
{
    \includegraphics[scale=.12]{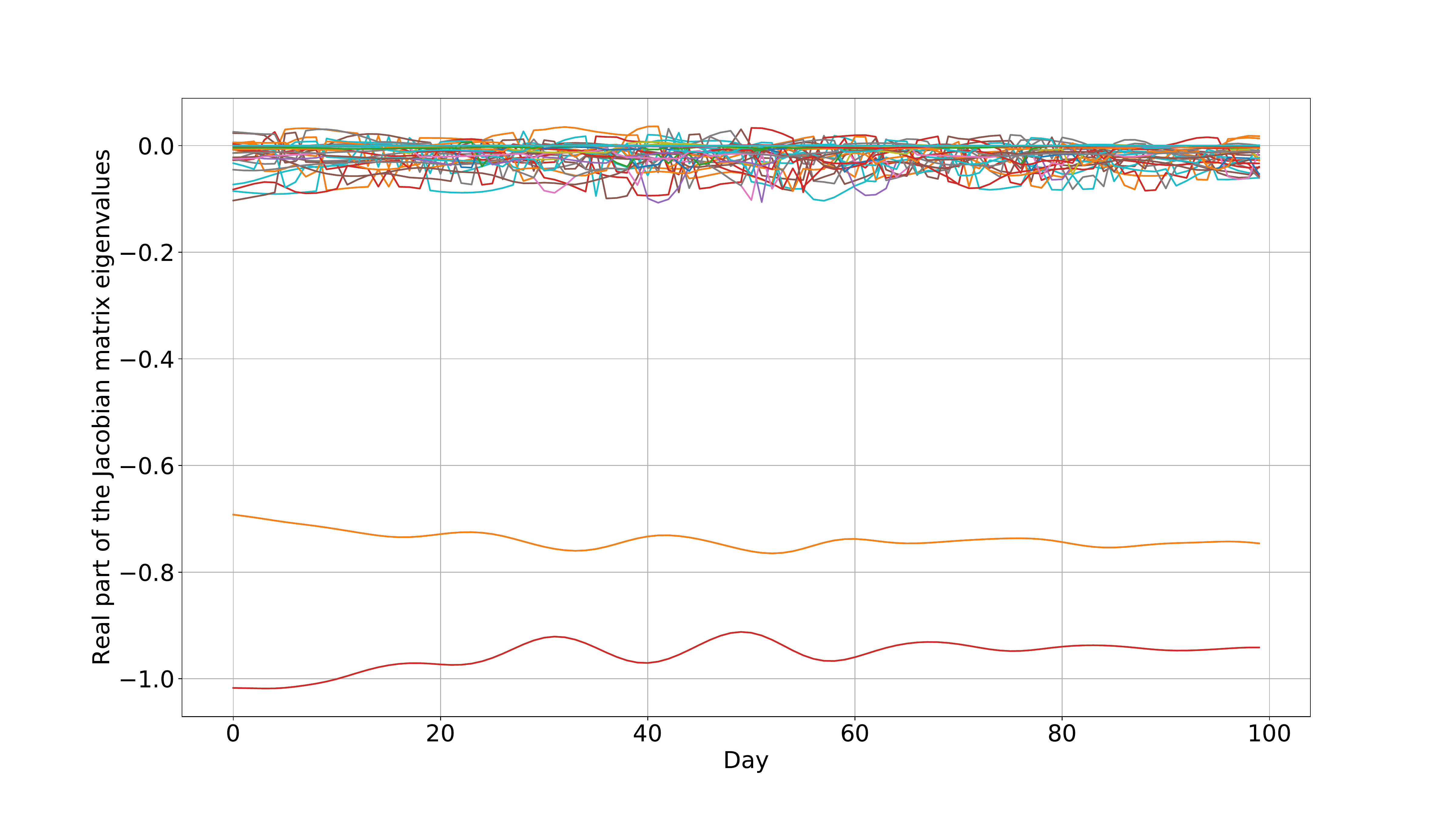}
    \label{fig:real_sla}
}
\subfloat[Eigenvalues imaginary part.]
{
    \includegraphics[scale=.12]{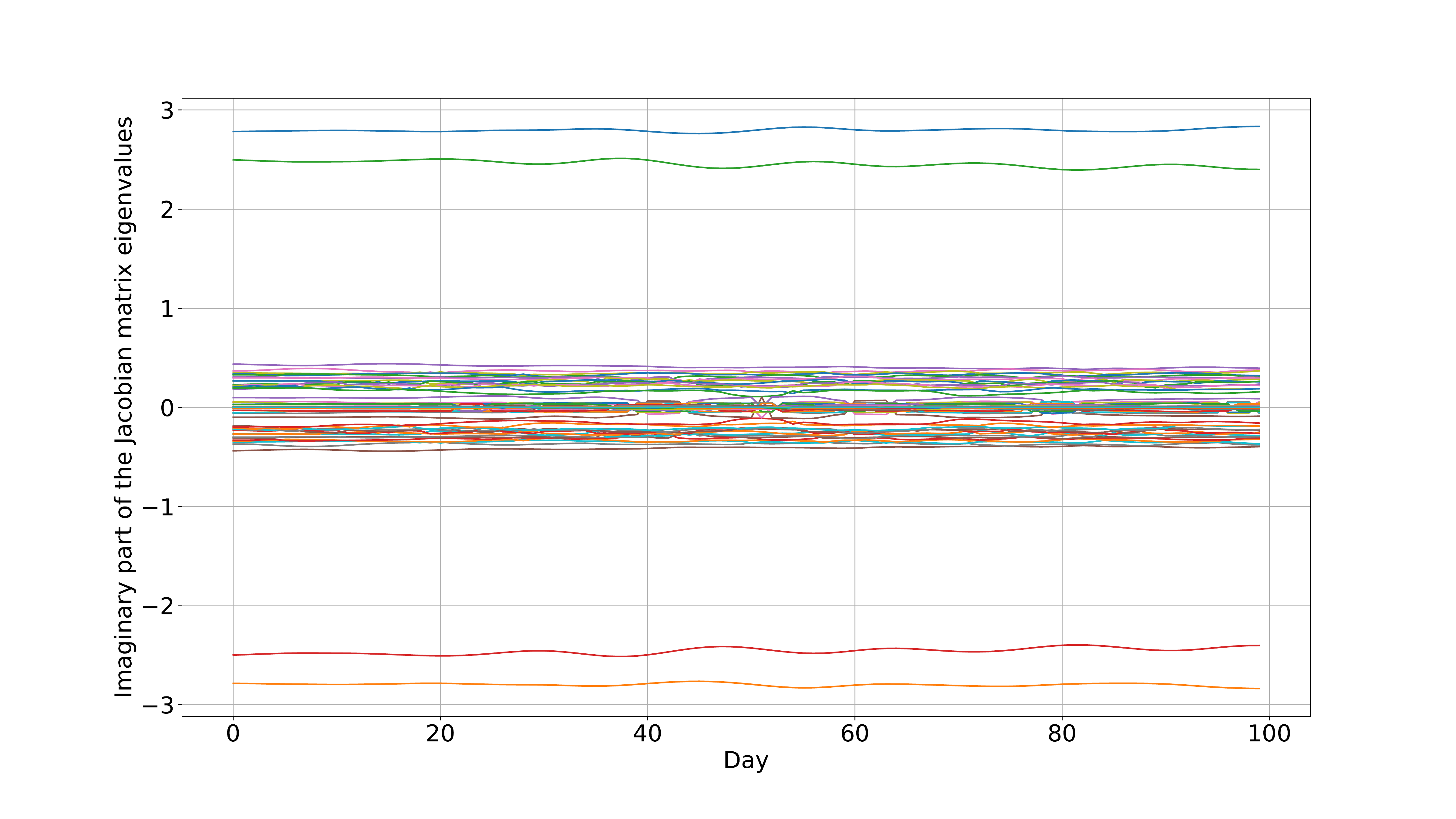}
    \label{fig:imag_sla}
}

\subfloat[Eigenvalues modulus.]{%
  \includegraphics[width=8cm,height=5cm]{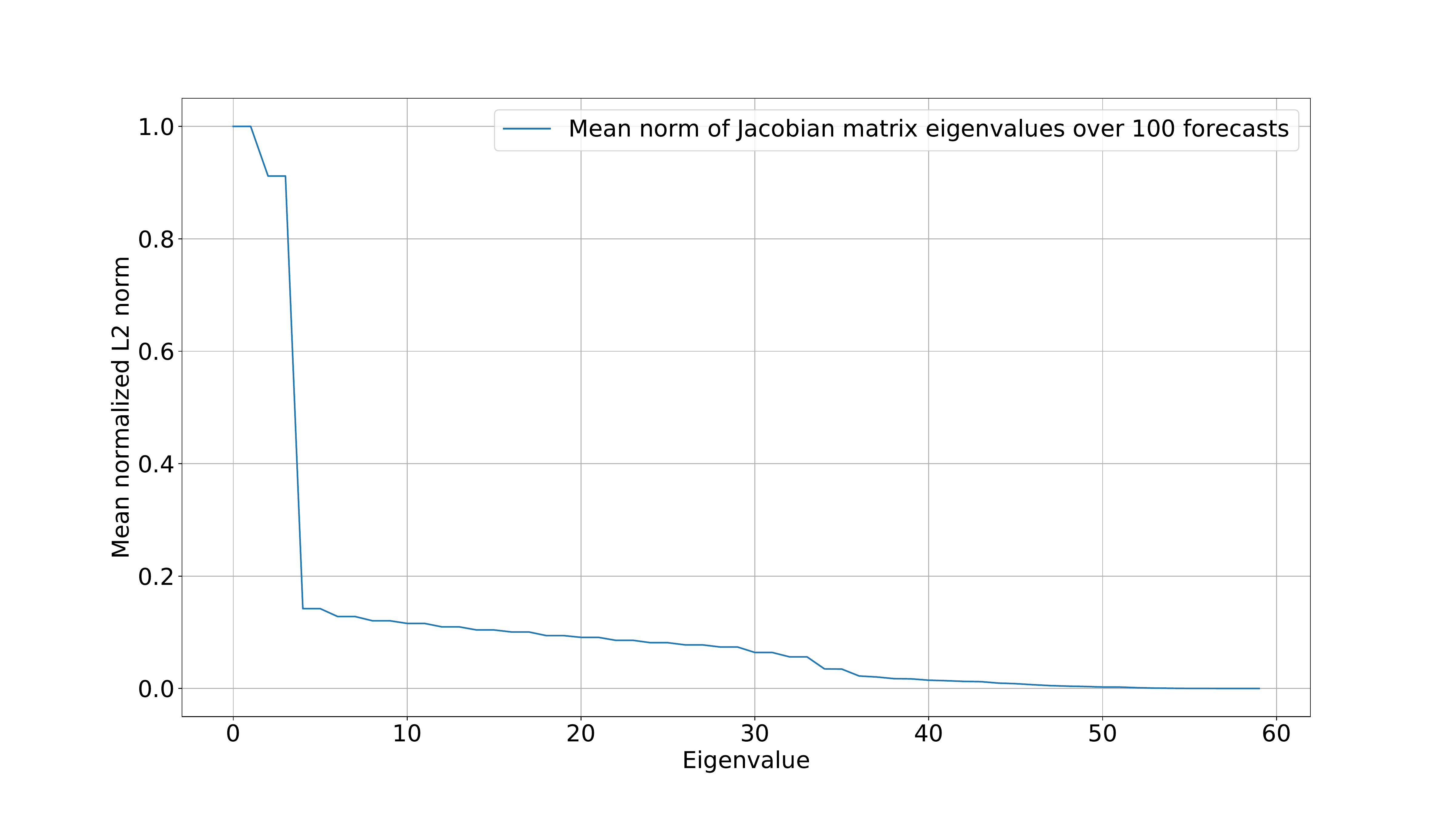}%
  \label{fig:mod_sla}
}

\caption{{{\bf   \em Analysis of the eigenvalues of the NbedDyn model Jacobian matrix.}: Sea Level Anomaly case-study with $d_E = 60$}}
\label{fig:sla_eig}
\end{figure}

The analysis of the eigenvalues of the Sea Level Anomaly model in the other hand are not as straightforward as in the case of the Lorenz model since we do not have any idea on the analytical form of the underlying dynamical model. Fig. \ref{fig:sla_eig} illustrates that using a 60 dimensional latent space for the NbedDyn model, only 50 eigenvalues have non-zero modulus and thus, are effectively influencing the underlying dynamics. The conclusion in this case is that the observed SLA data evolve in a 50 dimensional latent space parametrised by the dynamical model $f_\theta$.

\section{Additional figures of the Lorenz 63 experiment}
\label{sec:figs_Lorenz}
\begin{figure}[h]
\subfloat[NbedDyn forecasting.]{%
  \includegraphics[width=\columnwidth,height=2.5cm]{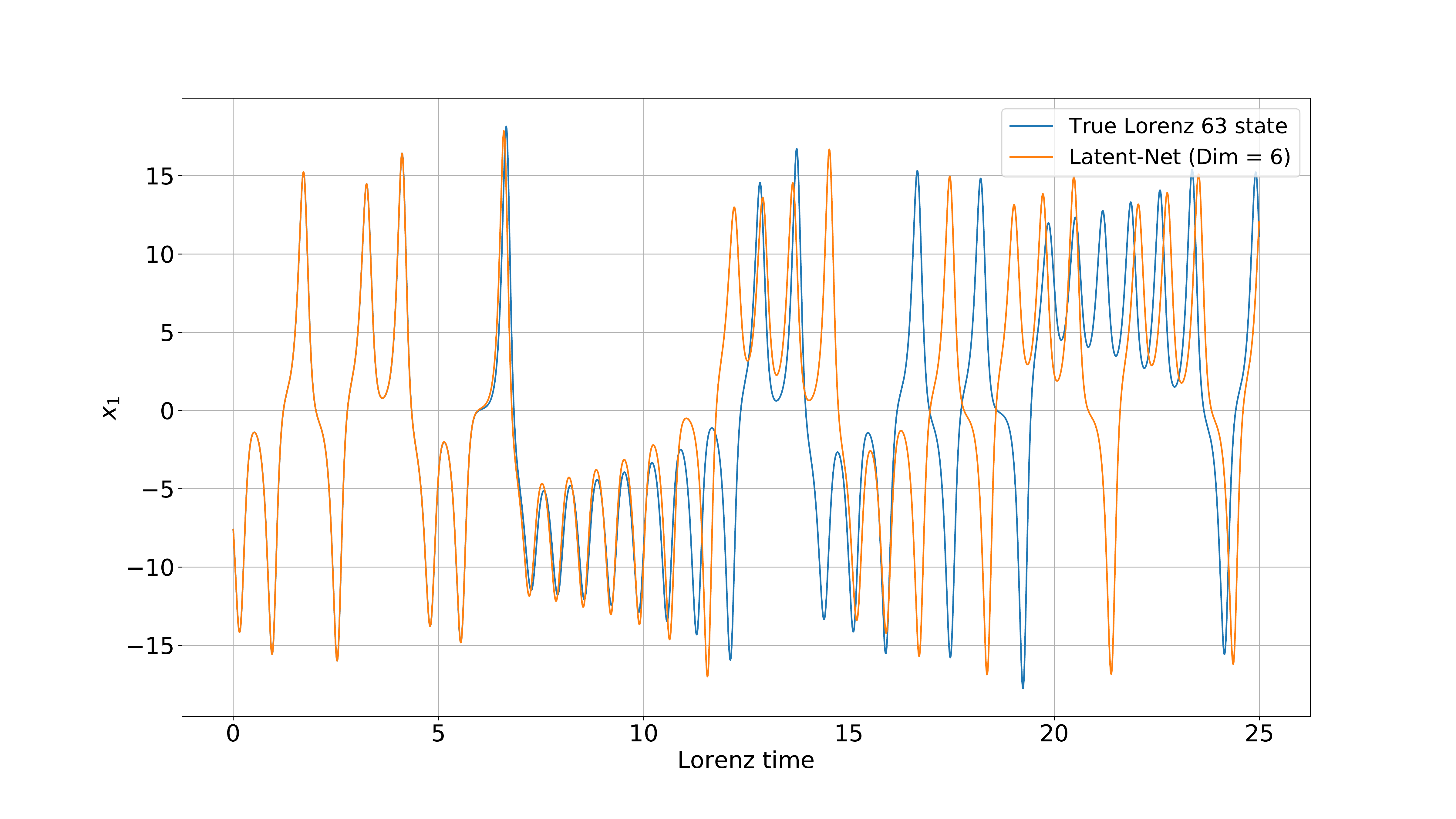}%
}

\subfloat[Analog Forecasting.]{%
  \includegraphics[width=\columnwidth,height=2.5cm]{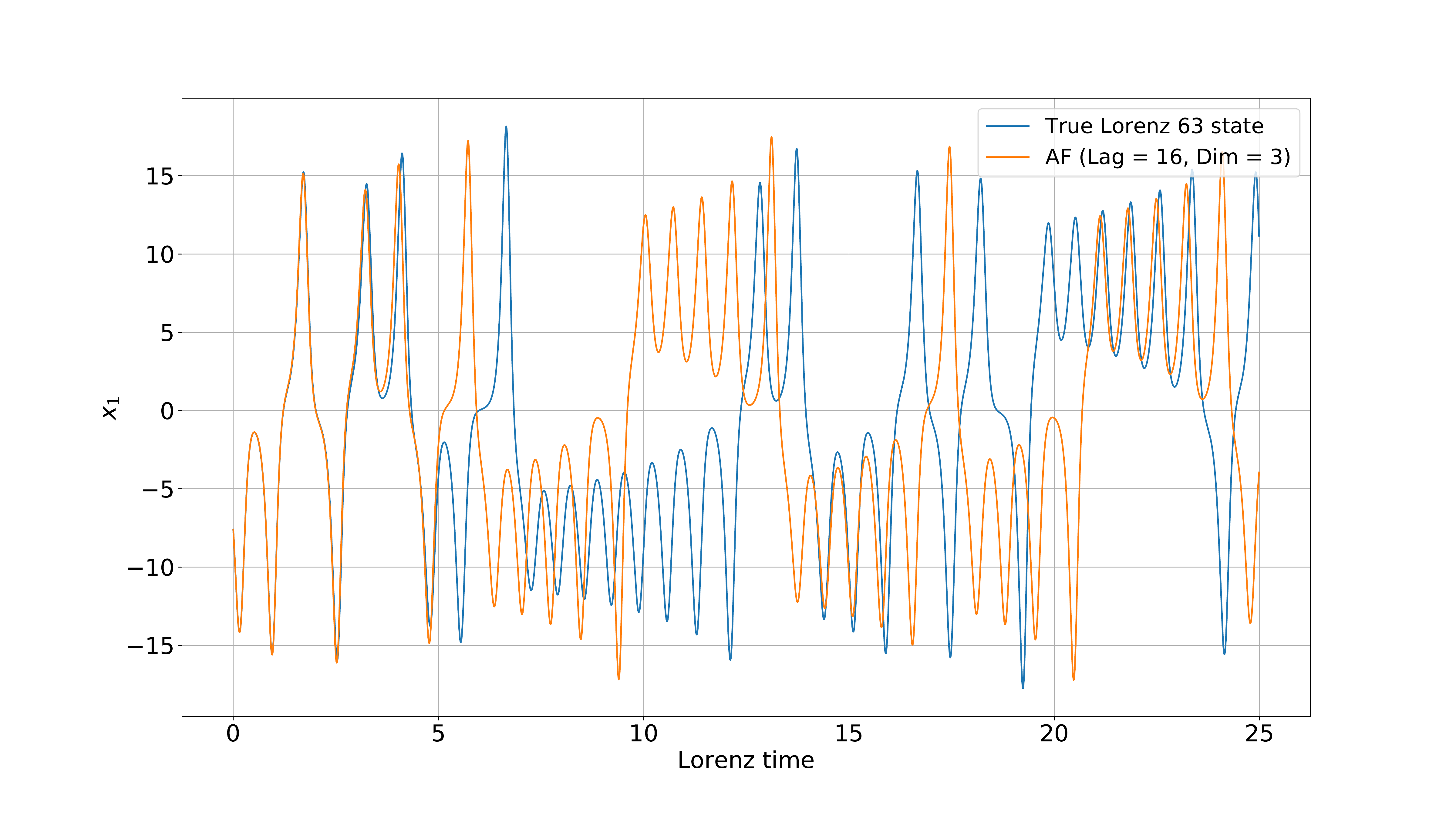}%
}

\subfloat[Sparse Regression forecasting.]{%
  \includegraphics[width=\columnwidth,height=2.5cm]{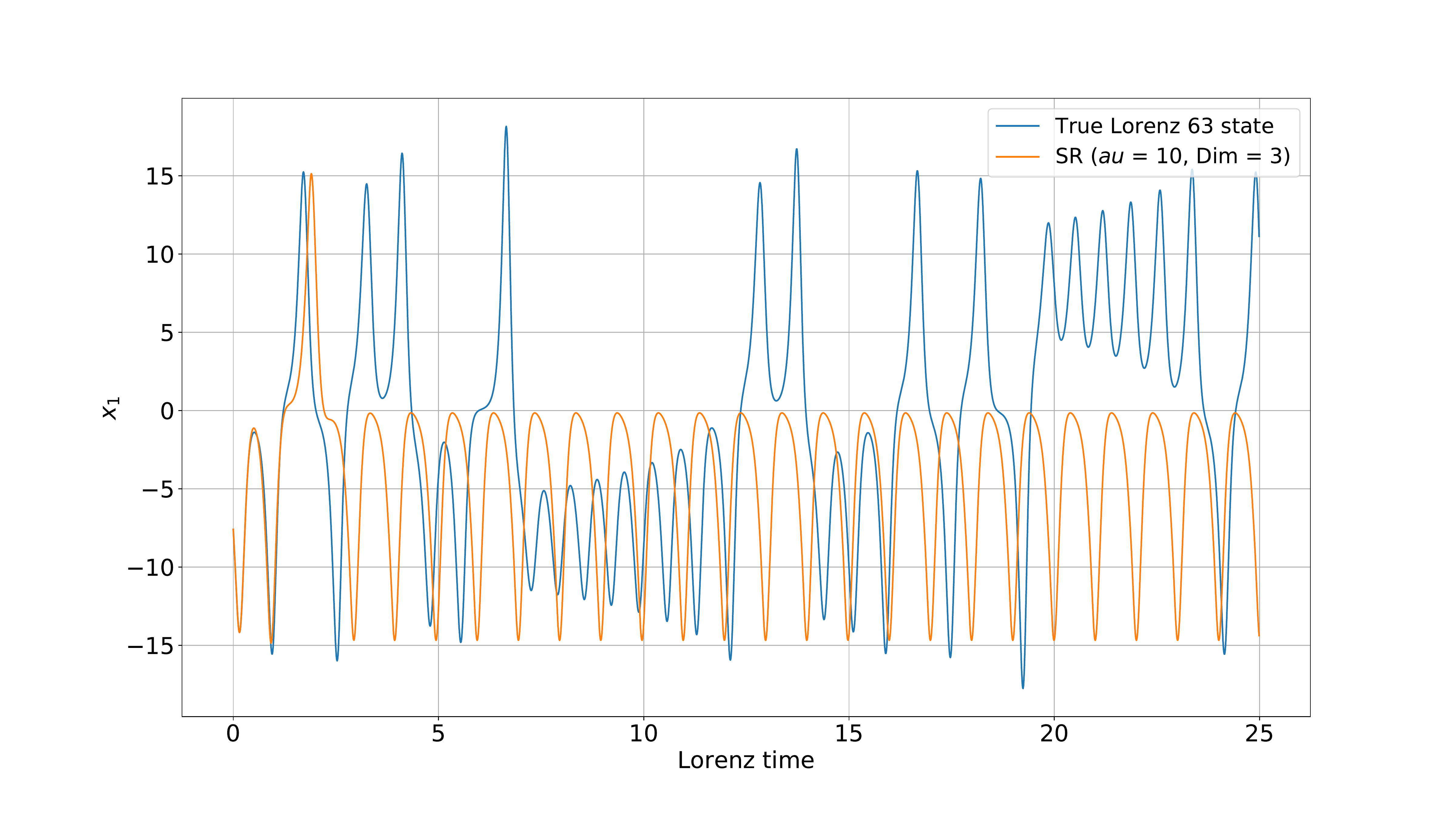}%
}

\subfloat[Latent-ODE forecasting.]{%
  \includegraphics[width=\columnwidth,height=2.5cm]{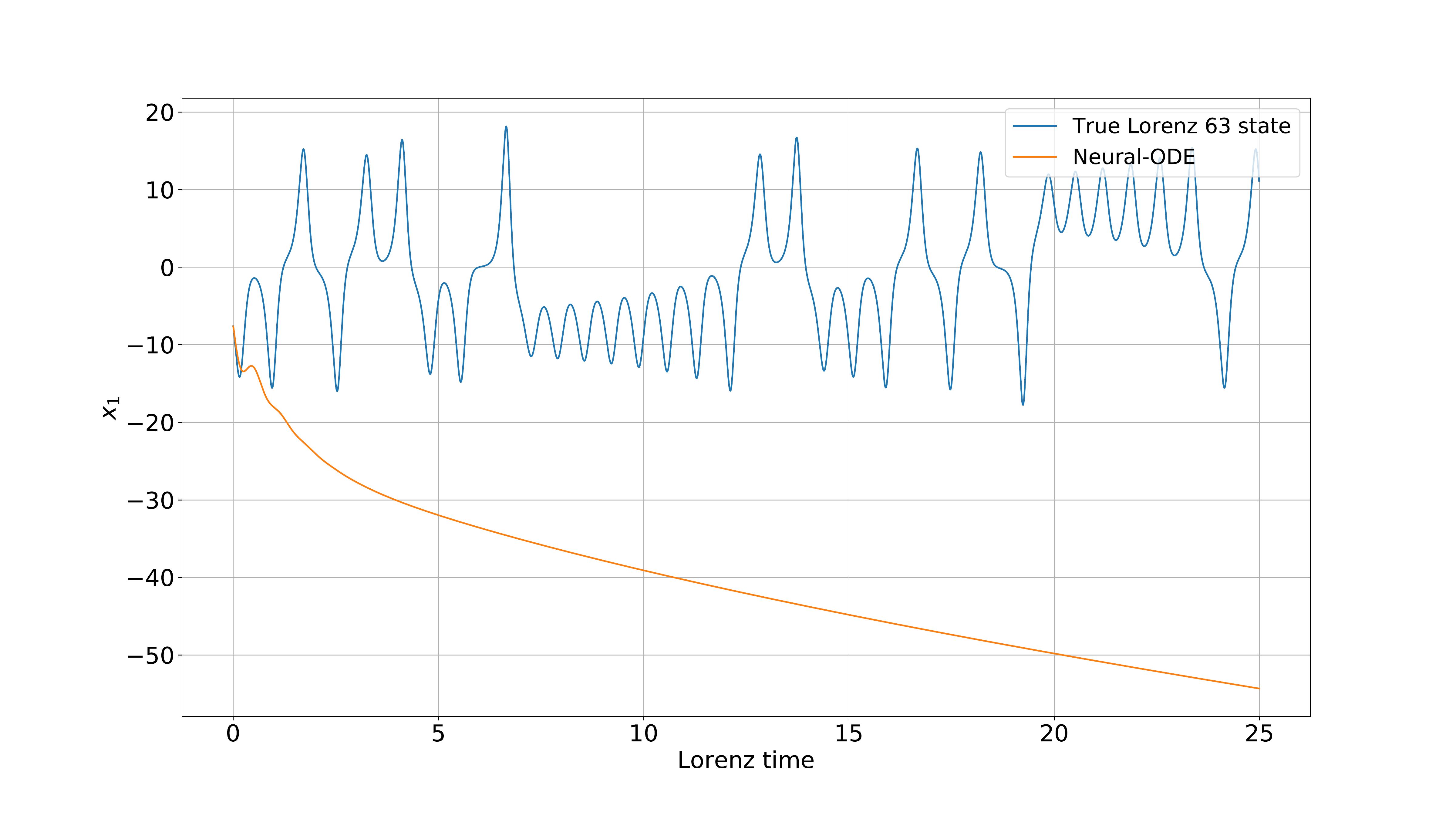}%
}

\caption{{{\bf   \em Generated time series of the proposed models.} : Given the same initial condition, we generated a time series of 2500 time steps.}
}
\label{fig:fore_lore}
\end{figure}

We illustrate the forecasting performance of the tested models for the Lorenz-63 experiment through an example of forecasted trajectories in Fig. \ref{fig:fore_lore}. Our model with $d_E=6$ leads to a trajectory similar to the true one up to $9$ Lyapunov times, when the best alternative approach diverge from the true trajectory beyond $5$ Lyapunov times.

\begin{figure}[hb]
\centering
\includegraphics[clip,width=8cm,height=4cm]{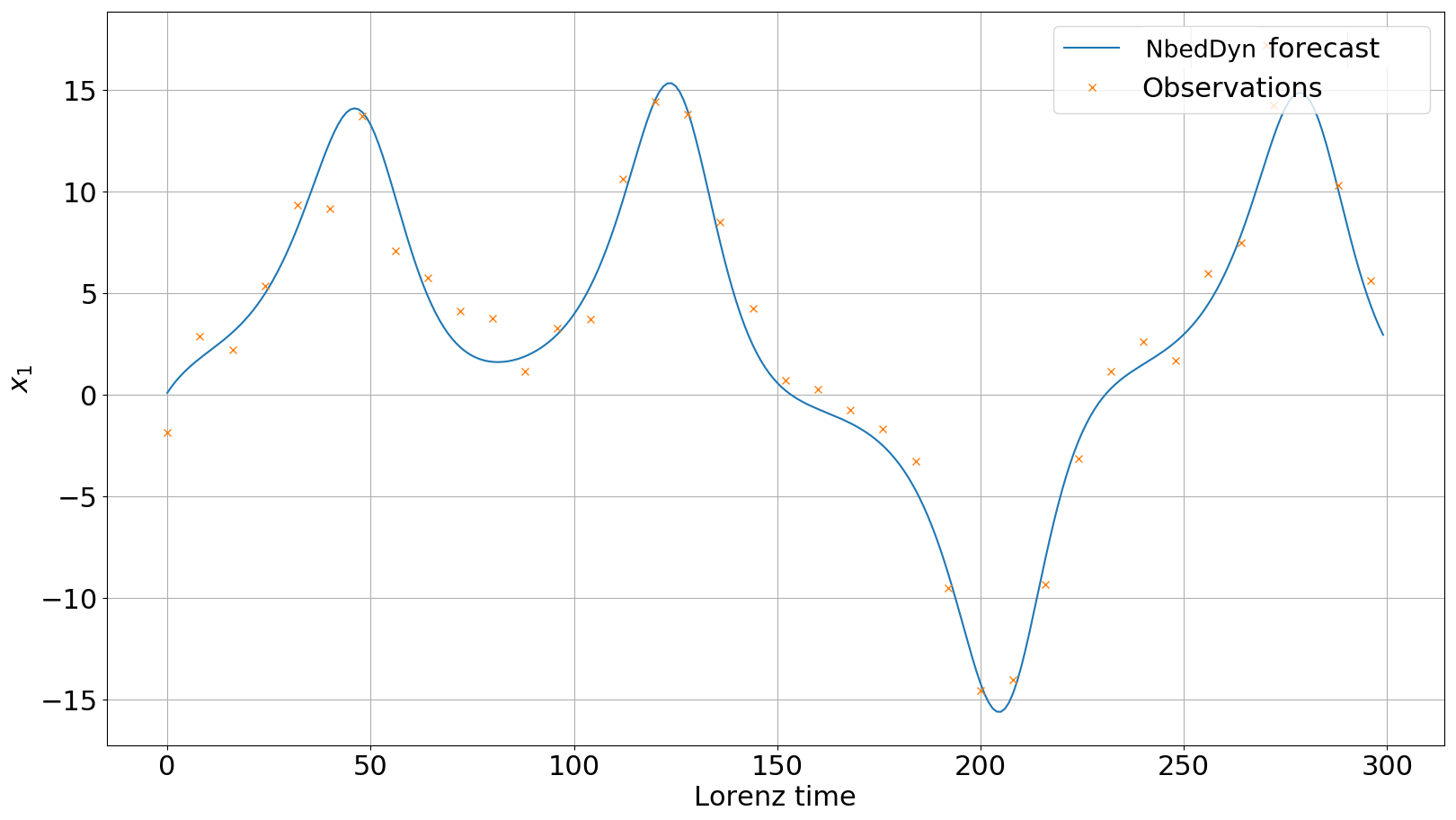}

     \caption{{{\bf   \em Forecasted Lorenz 63 state sequence given noisy and partial observations}:  Given noisy and partial observations, our model optimizes equation (7) to infer an initial condition that minimize the forecasting of the observations.}}
     \label{fig:reconst}
   \end{figure}

An other interesting experiment is to find the initial condition for new observation data. This issue is addressed as presented in section 3 as follow. Given a new noisy and partial observation sequence (Fig. \ref{fig:reconst}), we first look for a potential initial condition in the inferred training latent state sequence. This initial condition is then optimized using the cost function described by equation (7) to minimize the forecasting error of the new observation sequence.

\newpage

\section{Additional figures of the Sea level Anomaly experiment}
\label{sec:figs_SLA}

\begin{figure}[h]
\subfloat[Ground truth.]{%
  \includegraphics[width=\columnwidth,height=3cm]{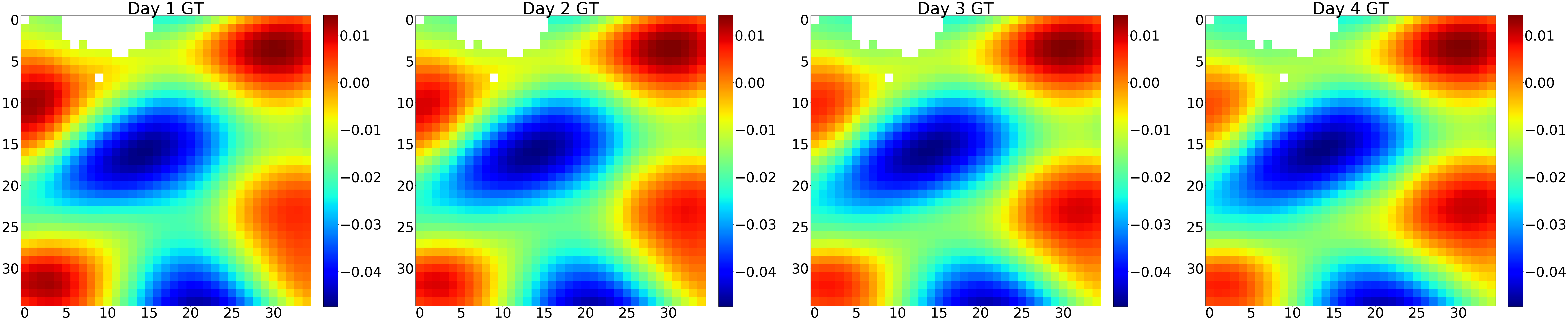}%
}

\subfloat[Analogs forecasting.]{%
  \includegraphics[width=\columnwidth,height=3cm]{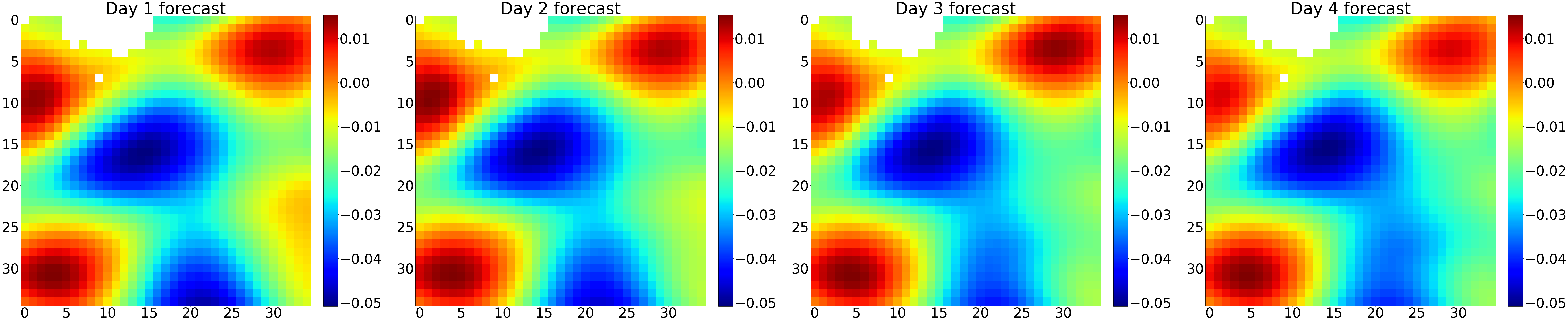}%
}


\subfloat[Latent-ODE forecast.]{%
  \includegraphics[width=\columnwidth,height=3cm]{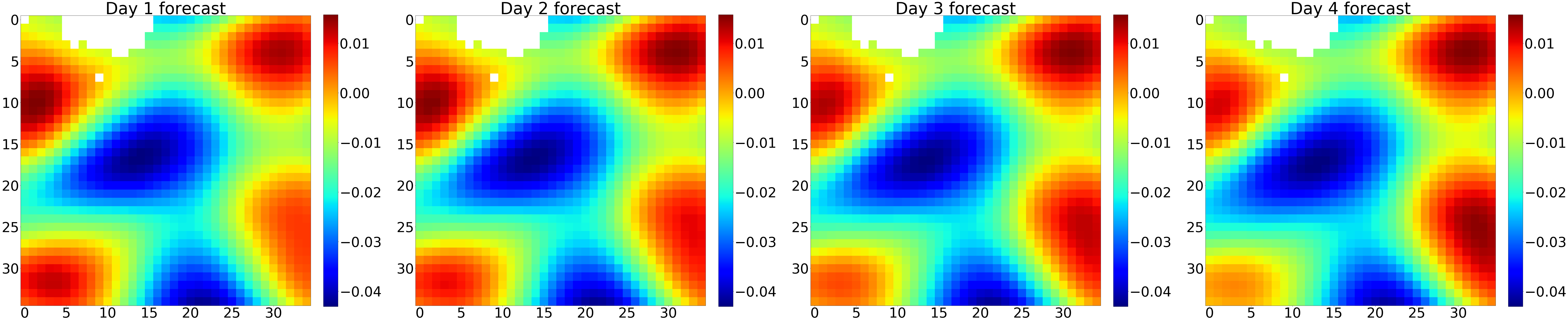}%
}

\subfloat[NbedDyn forecast.]{%
  \includegraphics[width=\columnwidth,height=3cm]{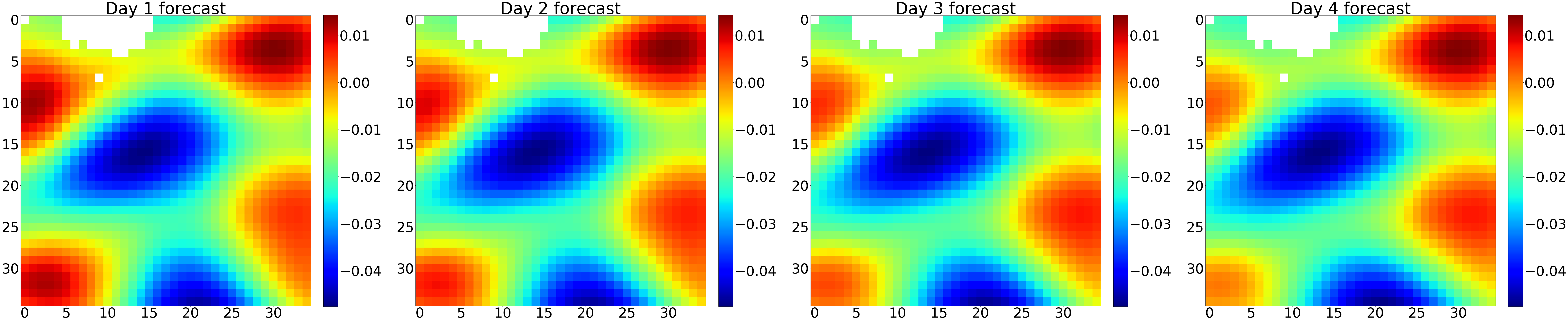}%
}

\caption{{{\bf   \em Forecasted SLA states of the proposed models.} }
}
\label{fig:fore_SLA}
\end{figure}

\begin{figure}[h]
\subfloat[Ground truth gradient.]{%
  \includegraphics[width=\columnwidth,height=3cm]{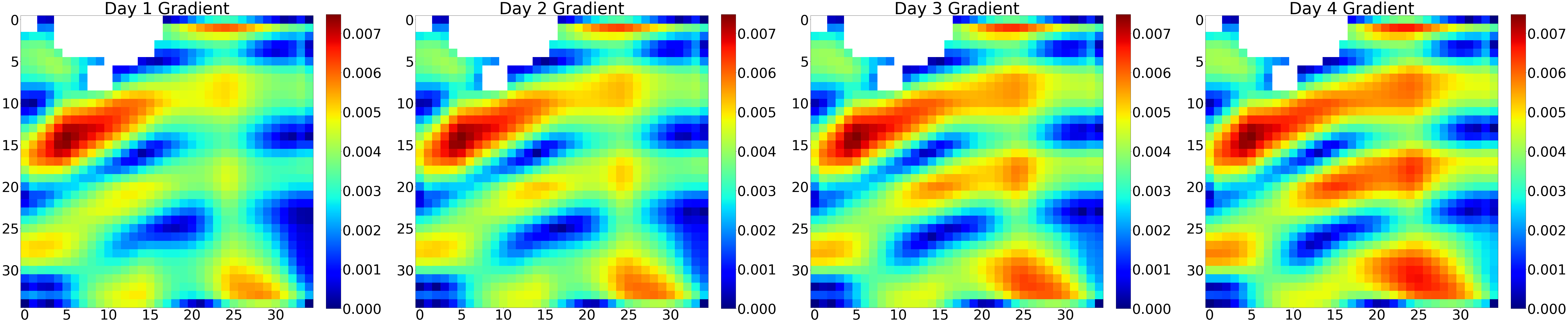}%
}

\subfloat[Analogs forecasting SLA gradient.]{%
  \includegraphics[width=\columnwidth,height=3cm]{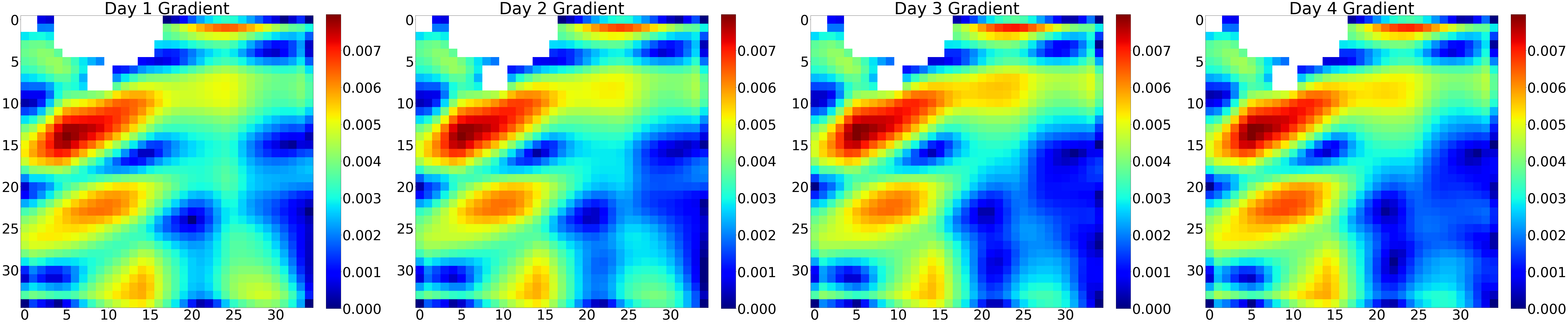}%
}


\subfloat[Latent-ODE SLA gradient.]{%
  \includegraphics[width=\columnwidth,height=3cm]{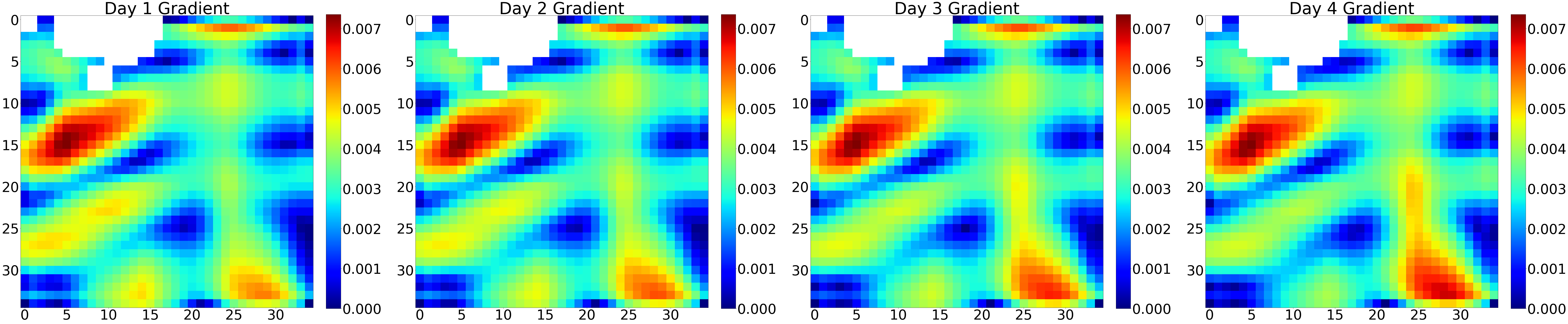}%
}

\subfloat[NbedDyn SLA gradient.]{%
  \includegraphics[width=\columnwidth,height=3cm]{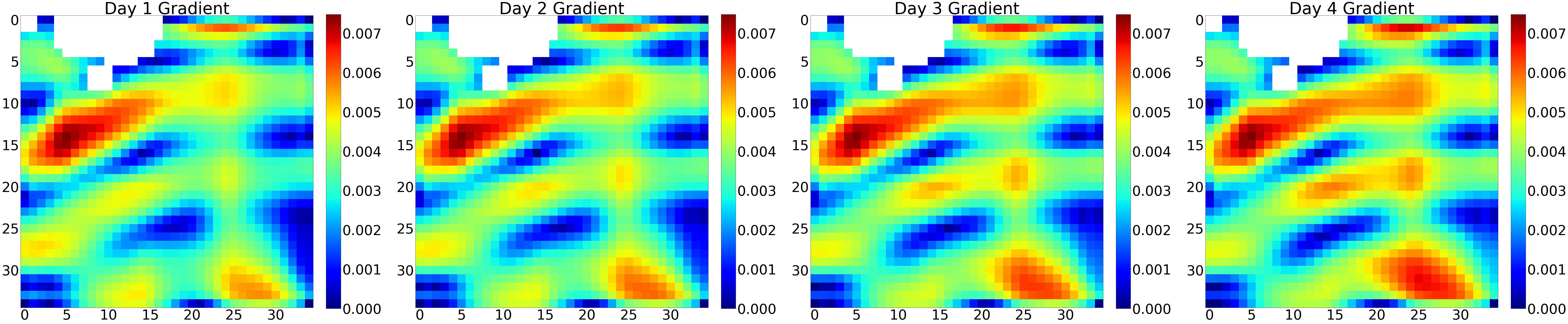}%
}

\caption{{{\bf   \em Gradients of the forecasted SLA states.} }
}
\label{fig:fore_Gradients}
\end{figure}

\begin{figure}[h]
\subfloat[Analog forecasting.]{%
  \includegraphics[width=\columnwidth,height=4cm]{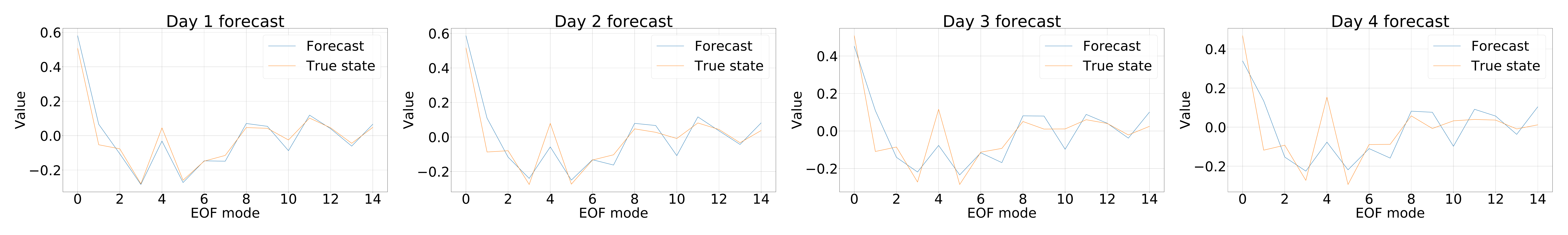}%
}

\subfloat[Latent-ODE.]{%
  \includegraphics[width=\columnwidth,height=4cm]{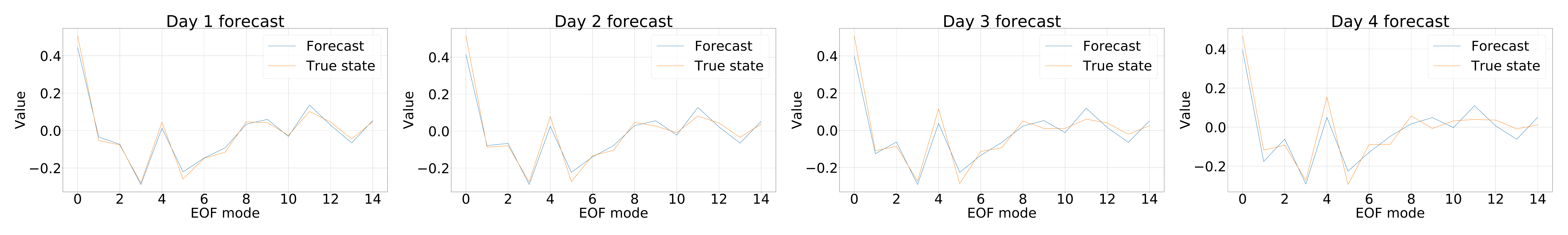}%
}

\subfloat[NbedDyn.]{%
  \includegraphics[width=\columnwidth,height=4cm]{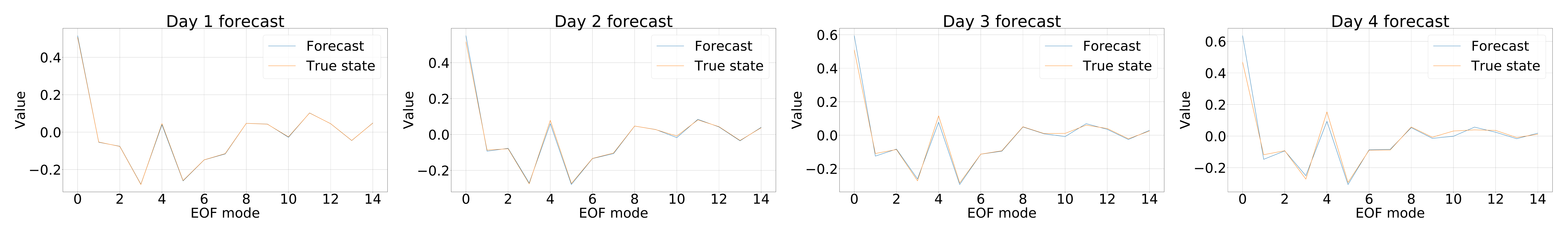}%
}

\caption{{{\bf   \em Forecasted EOF components of the proposed models.} }
}
\label{fig:fore_EOF}
\end{figure}

Forecasted states of the Sea Level Anomaly and their gradients are illustrated in Fig. \ref{fig:fore_SLA}, \ref{fig:fore_Gradients} and \ref{fig:fore_EOF}. The visual analysis of the forecasted SLA states emphasize the relevance of the proposed NbedDyn model. While state of the art approaches generally overestimate the time evolution of some structures such as eddies, our model is the only one to give near perfect forecasting up to 4 days.

\end{document}